\documentclass[logo, 11pt, a4paper]{deepmind}
\pdfoutput=1

\usepackage{parskip}        
\usepackage{graphicx}
\usepackage{amsmath}
\usepackage{caption}
\usepackage{subcaption}
\usepackage{float}
\usepackage{xcolor}
\usepackage{hyperref}

\usepackage{listings}
\usepackage{color}
\definecolor{codegreen}{rgb}{0,0.5,0}
\definecolor{codeblue}{rgb}{0,0,0.9}
\definecolor{codegray}{rgb}{0.5,0.5,0.5}
\definecolor{codepurple}{rgb}{0.58,0,0.82}
\definecolor{backcolour}{rgb}{0.95,0.95,0.92}
\definecolor{backcolour2}{rgb}{0.9,0.9,0.9}
\definecolor{codered}{rgb}{0.5,0,0}
\definecolor{textcodered}{rgb}{0.4,0,0}
\definecolor{palegray}{rgb}{0.98,0.98,0.99}

\lstdefinestyle{mystyle}{
    backgroundcolor=\color{backcolour},   
    commentstyle=\color{codered},
    keywordstyle=\color{codeblue},
    numberstyle=\tiny\color{codegray},
    stringstyle=\color{codegreen},
    breakatwhitespace=false,         
    breaklines=true,                 
    captionpos=b,                    
    keepspaces=true,                 
    numbersep=5pt,                  
    showspaces=false,                
    showstringspaces=false,
    showtabs=false,                  
    tabsize=2,
    otherkeywords={with},
    basicstyle=\ttfamily\footnotesize
}

\usepackage{enumitem}
\setitemize{noitemsep,topsep=0pt,parsep=0pt,partopsep=0pt}

\title{Semi-analytical Industrial Cooling System Model for Reinforcement Learning}

\correspondingauthor{cyuri@deepmind.com, praneetdutta@deepmind.com}

\author[*,1]{Yuri Chervonyi}
\author[*,1]{Praneet Dutta}
\author[2]{Piotr Trochim}
\author[1]{Octavian Voicu}
\author[1]{Cosmin Paduraru}
\author[2, 3]{Crystal Qian}
\author[1]{Emre Karagozler}
\author[1]{Jared Quincy Davis}
\author[4]{Richard Chippendale}
\author[3]{Gautam Bajaj}
\author[1]{Sims Witherspoon}
\author[1]{Jerry Luo}

\affil[*]{Equal contributions}
\affil[1]{DeepMind}
\affil[2]{Work done at DeepMind}
\affil[3]{Google}
\affil[4]{Work done at COMSOL}

\begin{abstract}
We present a hybrid industrial cooling system model that embeds analytical solutions within a multi-physics simulation. This model is designed for reinforcement learning (RL) applications and balances simplicity with simulation fidelity and interpretability. The model’s fidelity is evaluated against real world data from a large scale cooling system. This is followed by a case study illustrating how the model can be used for RL research. For this, we develop an industrial task suite that allows specifying different problem settings and levels of complexity, and use it to evaluate the performance of different RL algorithms.
\end{abstract}

\begin{document}
\maketitle


\section{Introduction}

\subsection{Background and Motivation} 

Industrial systems account for 54\% of global energy usage \cite{industrial-sector-energy-consumption} and 24\% of global net anthropogenic Greenhouse Gas (GHG) emissions. The latter percentage rises to 34\% if indirect emissions from energy are included, which would make industrial systems the highest emitting sector \cite{portner2022climate}. Due to increasing global demand for the products and services enabled by industrial systems, emissions from this sector will continue to rise \cite{krausmann2018resource}. However, there is strong evidence that interventions such as reduction in energy use per unit of output \cite{wang2019evolution}, lightweight designs and extended product lifetimes can facilitate critical emissions reductions across industrial systems \cite{hertwich2019material}. Yet, optimizing industrial systems is not straightforward; subsectors such as metals, chemicals, waste and cement require customized approaches accounting for different materials, processes and facility configurations. 

Recent work has shown that reinforcement learning can be leveraged to efficiently control and optimize industrial processes (e.g. see \cite{tortorelli2022parallel, zhan2021deepthermal}). Among these processes, \textit{cooling}/heating has been the main focus for many applied researchers due its wide usage in residential and commercial buildings, automobiles, industrial plant machinery, nuclear reactors, and many other types of machinery. Examples of applying RL in real cooling/heating systems include reducing energy consumption for cooling Google data centers by 40\%  \cite{google-ai-data-center} and controlling the HVAC systems, hot water tank and heat pumps in residential \cite{Kazmi_2018} and industrial facilities \cite{chen2019gnu, zhang2019whole, hanumaiah2021distributed}.

However, applying RL in live facilities comes with technical challenges, such as:
\begin{itemize}
    \item \textit{Limited exploration}: During exploration, the agent must act cautiously to avoid real-world consequences such as overheating a critical facility. The agent cannot explore all combinations of the state and action space, resulting in low-variance data. Furthermore, explore-exploit schedules must be negotiated with the facility, often with the motivation for the agent to exploit sooner rather than later.
    \item \textit{Generalization}: Although many industrial facilities are similar, policies trained on one facility may not generalize to another facility due to slight differences in the input feature space. Solutions are not robust to different equipment conditions, weather patterns, and other sources of variance, as training is limited to experiences with a singular system during a fixed amount of real time interactions.
    \item \textit{Sample efficiency}: In live facilities, the agent interacts with the environment and receives information at coarse frequencies such as 5 minutes or an hour; this limits the amount of data that can be used for training and also slows the process of validating existing models. From an algorithmic perspective, we are constrained to approaching this problem with sample-efficient algorithms.
    \item \textit{Infrastructure and stakeholder management}: Interacting with a live facility requires significant additional engineering effort, including developing a shared data schematic and data transfer API with the facility partner, writing custom logic to integrate facility-specific features, and ensuring reliability and availability during deployment. 
    \end{itemize}
    
With these challenges in mind, we present a novel semi-analytical simulation model of an industrial cooling facility, developed specifically for reinforcement learning applications. Interacting with a simulator rather than a live facility greatly increases the amount of available data and allows the agent to explore a broader action space safely. We provide a set of analytical equations as well as the description of how to embed them into a multi-physics simulation software. We hope that this will be a useful blueprint for practitioners who want to improve industrial cooling system efficiency using RL.

In order to allow us to create agents that generalize well, we also provide the high level design for an Industrial Task Suite (ITS) that allows us to specify different RL tasks based on the simulator. ITS can be used to generate a curriculum of scenarios and defines highly parameterized facility configurations to enable flexible experimentation and the development of generalizable agents.

\subsection{Related Work}

\subsubsection{Simulation} 
There are several tools offered for developing simulations for optimizing industrial systems and processes. Among such tools, EnergyPlus \cite{crawley2001energyplus} is a popular open-source library for whole-building energy simulations. It is used by by many researchers, especially for the RL applications \cite{chen2018optimal, moriyama2018reinforcement, hanumaiah2021distributed}. Although it provides full-featured functionality, there are fidelity limitations that can be crucial when simulating real-world environments \cite{kim2011difficulties, mun2020limitations}. 

\subsubsection{Optimization Research}
Work in the domain of simulation-enabled industrial system optimization research can broadly be segmented into those primarily addressing questions pertaining to data, infrastructure, and methods. Nweye et al. \cite{nagy2021real} has probed challenges around safety-critical building control and offline training, arguing that the sequence of operations policy used to generate offline trajectory data has a dramatic effect on the downstream performance of the associated offline RL agents. With regards to infrastructure, \cite{hanumaiah2021distributed} proposes a scheme for weaving together popular free and open source tools to create common building control research infrastructure and environments. It illustrates a scheme for framing HVAC control as a Deep RL problem using OpenAI gym \cite{brockman2016openai}, EnergyPlus \cite{crawley2001energyplus}, and Ray RLlib \cite{liang2018rllib}. The authors overlay real-world historical weather data from Toronto, Canada, and simulate a year's worth of weather dynamics, producing ablations to show the impact of weather, and energy-temperature penalty coefficients. In terms of the methods, many approaches have been proposed for energy savings and temperature
control. For example, Gao et al. \cite{gao2019energy} explored the use of a Deep Deterministic Policy Gradients algorithm (DDPG) \cite{lillicrap2015continuous}; Zhang et al. \cite{zhang2019whole} considered the Asynchronous Advantage Actor Critic (A3C) \cite{mnih2016asynchronous}. Other adjacent works explore applications of deep RL in other industrial control constrained optimization settings, such as assembly line control \cite{tortorelli2022parallel} and combustion control \cite{zhan2021deepthermal}, proposing parallel agent training schemes to optimize throughput and mitigate simulation step-time limitations.

\subsection{Summary of Results}

Our main contributions include:
\begin{itemize}
    \item Equations and design for our \textit{Semi-analytical Industrial Cooling System Model}, a simulation of a large cooling facility that combines analytical solutions with a multi-physics simulation software (Section \ref{sec:semi-analytical-simulation-model}). The model's fidelity is verified against real-world data from a large commercial cooling facility in the United States (Section \ref{sec:simulation-fidelity}).
    \item Design for the \textit{Industrial Task Suite} (ITS), a framework for RL research in the domain of industrial systems (Section \ref{sec:its}). Although we primary use it for the industrial cooling applications, several other use-cases are presented in Appendix \ref{sec:its-other-applications}. ITS is designed to allow more granular configurability of environments compared to the available task suites (such as the DeepMind Control Suite \cite{tassa2018deepmind}); specifically, allowing the combination of independently-defined noise, constraints, scenarios, tasks and simulators for modelling industrial problems.
    \item \textit{Tasks} for industrial cooling systems categorized by their perceived complexity (Section \ref{sec:HVAC-tasks}). We also show how to define other new tasks using the capabilities of the ITS.
    \item \textit{RL Agents Benchmarks} on these tasks (Section \ref{sec:experiments-and-results}).
\end{itemize}

\section{Industrial Cooling System Simulation}
\label{sec:simulation}

\subsection{Chilled Water Plant System Overview}
\label{sec:chilled-water-plant-overview}

A cooling system is an apparatus employed to keep the temperature of a structure or device from exceeding limits imposed by needs of safety and efficiency \cite{cooling-system-britannica}. Cooling systems use air or a liquid as a medium to absorb heat. Many applications use water because it has a high boiling point and specific heat. In the following sections, we focus on chilled water plants, cooling systems that leverage water-cooled pipelines.

\begin{figure*}[h]
    \includegraphics[width=\textwidth]{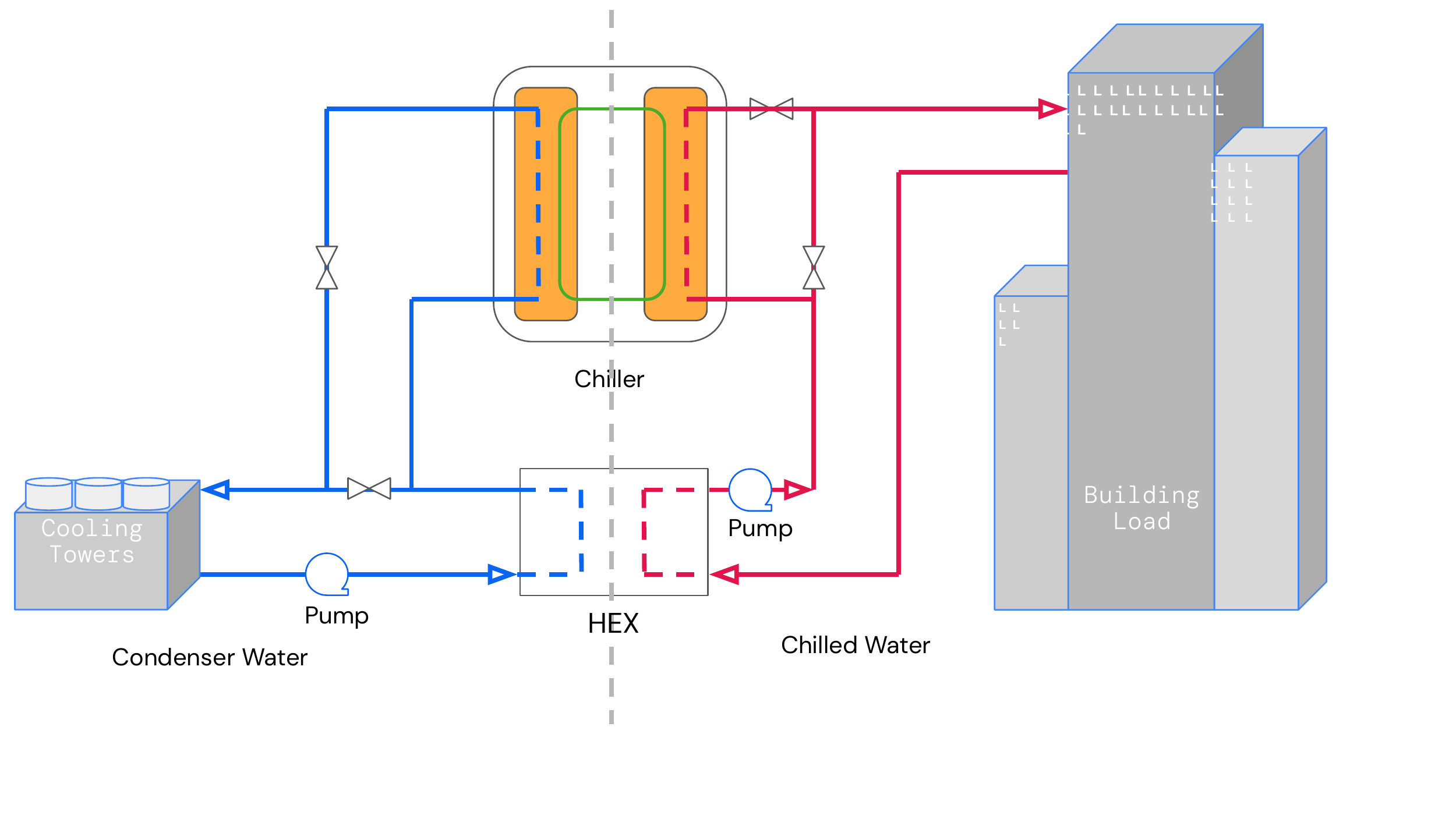}
    \caption{A high level overview of a typical chilled water plant system. The main components include one or several chillers, a heat exchanger (HEX), pumps, cooling towers. Building load is heat transferred through the building envelope (walls, roof, floor, windows, doors etc.) and heat generated by occupants, equipment, and lights.}
    \label{fig:chilled-water-plant-overview}
\end{figure*}

A chilled water plant system is depicted on Figure \ref{fig:chilled-water-plant-overview}. These systems are typically divided into two broader halves. The chilled water side extracts heat from the building and is responsible for cooling the rooms, keeping them within the desired temperature range. The condenser water side interacts with the outside environment and, in turn, extracts heat from the chilled water side. The flow of water is regulated through pumps on both sides. No water mass is exchanged between the loops, being restricted to heat transfer via conduction and convection. The cooling process can be achieved in the cooling tower by the latent heat of evaporation and the chillers via the vapor compression cycle.

Chilled water plant systems use evaporative cooling, in which highly energetic water molecules are evaporated from the surface, resulting in reduced temperature for the remaining water \cite{hanif2022renewable}. There are two types of temperature that are  important in these cooling systems: \textit{dry bulb temperature} and \textit{wet bulb temperature}. Dry bulb temperature is measured by a regular thermometer, by exposing it to the stream of air while shielding it from radiation and moisture. Wet bulb temperature is the lowest possible temperature of water that can be achieved just by the evaporation of water. This can be measured by a thermometer covered in the water-soaked cloth (wet-bulb thermometer), over which air is passed. Dry bulb temperature and wet bulb temperature can be used to calculate the relative humidity of air.

\begin{figure*}[h!]
    \includegraphics[width=\textwidth]{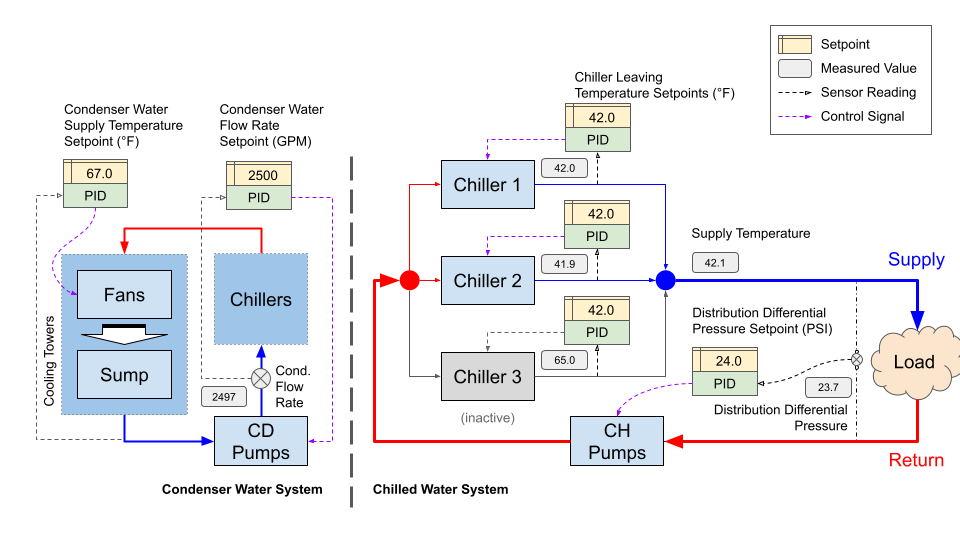}
    \caption{The detailed overview of a chilled water plant system, including the most important setpoints that can be controlled to achieve the required temperature. Here we see multiple chillers arranged in parallel to handle various external conditions. Unlike a system presented on Figure \ref{fig:chilled-water-plant-overview}, this design does not include a heat exchanger.
    }
    \label{fig:chilled-water-plant-overview-detailed}
\end{figure*}

The rate of cooling is controlled by setting various system parameters known as \textit{setpoints}, such as temperature, pressure, fluid flow rates etc. Industrial cooling systems are typically controlled by a Building Management System (BMS), a heuristic-based policy that operates on these setpoints to keep the building at a safe and comfortable desired temperature. A low level proportional–integral–derivative (PID) controller changes an actuator parameter to minimize the delta between this desired setpoint recommendation and a measured sensor reading (see Figure \ref{fig:chilled-water-plant-overview-detailed}). This setup, controlling setpoints for multi-objective optimization subject to constraints, lends itself well to reinforcement-learning approaches.

\subsection{Semi-Analytical Simulation Model}
\label{sec:semi-analytical-simulation-model}

We leverage a multi-physics simulator software (COMSOL) for modeling fluid and heat transfer across the system\footnote{This setup could be replicated by any applicable fluid dynamics simulator}. The main components of our cooling system are (see Figure \ref{fig:chilled-water-plant-overview}): 
\begin{itemize}
    \item Chillers.
    \item Cooling Towers.
    \item Chilled Water Pump Bank.
    \item Condenser Water Pump Bank.
    \item Air Handling Unit.
    \item Heat Exchanger.
    \item PID controllers.
    \item Valves, bends, friction loss parameters of the systems.
\end{itemize}

Our design goal is to balance simplicity with simulation fidelity and interpretability. We leverage a simplified pipe-flow module for modeling the fluid dynamics, which turns the meshed finite element analysis into a 2D curve, improving computational efficiency for 3D pipes with a finite diameter. We aim to model a large-scale industrial system where this approximation would be a reasonable assumption. Specifically, the pipe length to diameter ratio is large enough for the flow inside each pipe segment to be considered fully-developed. We utilize additional libraries for modeling the heat exchanger and PID controls, which controls components such as the temperature (via the compressor), pressure and flow rate (via the pumps) across the fluid loops.
External weather conditions such as the dry bulb temperature, wet bulb temperature, relative humidity and the building load profile form a critical aspect of the system; we simulate these using public historical weather databases.
Finally, chillers, cooling towers and pump banks are described via analytical solutions that have been verified using data from a real-world facility. Next, we discuss these solutions in detail.

\paragraph{Chiller and the improved Gordon-Ng model}

\begin{figure*}[h]
    \includegraphics[width=\textwidth]{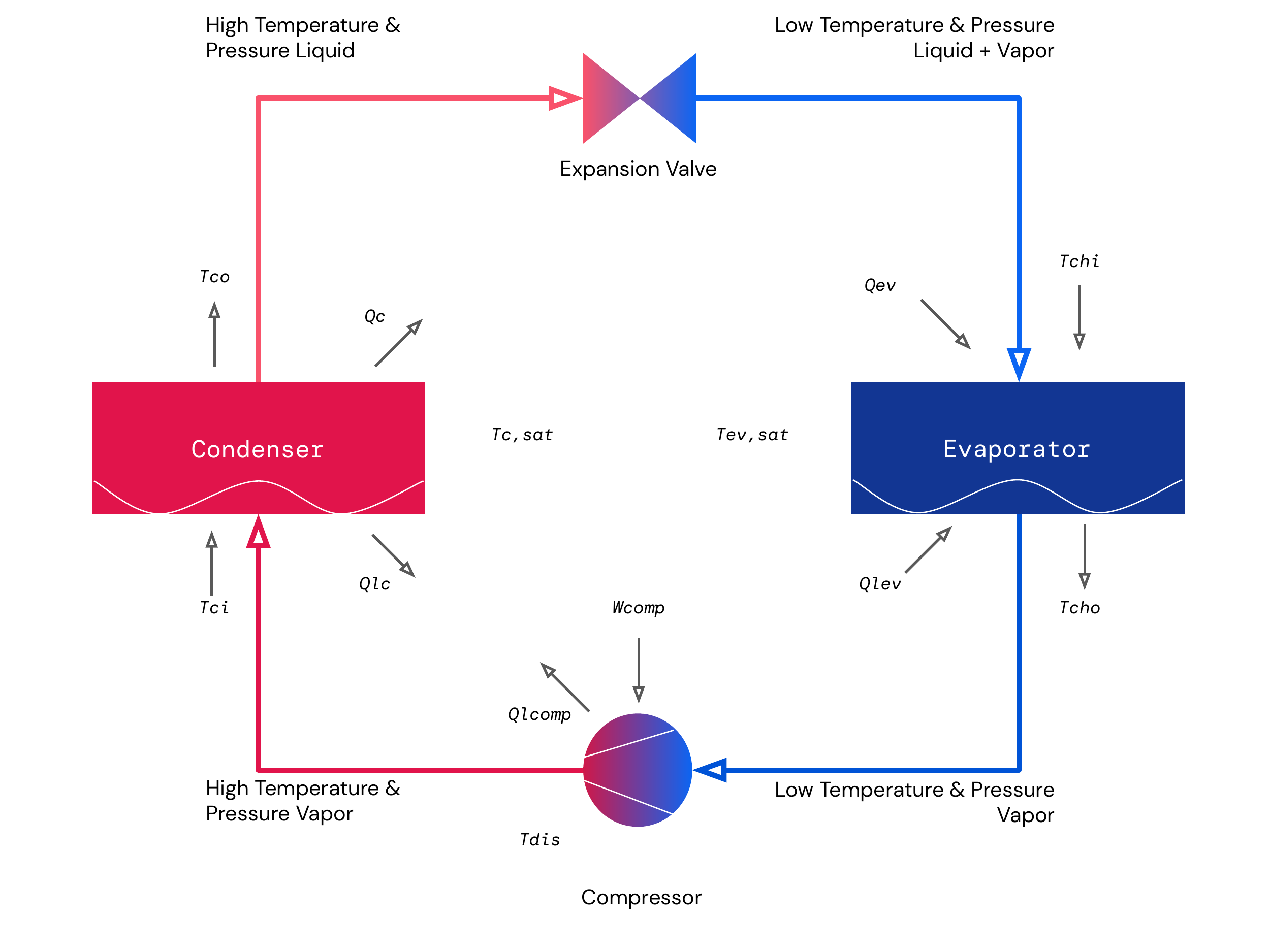}
    \caption{Vapor-compression cycle scheme used by a chiller. $T$s denote temperatures and $Q$s denote heat. For the details see the description under Eq. \ref{eq:chiller-4}. This diagram is adapted from \cite{foliaco2020improving}.}
    \label{fig:chiller}
\end{figure*}

For the chiller model we start with \cite{foliaco2020improving}, where the canonical Gordon-Ng model \cite{gordon1995centrifugal} was simplified and improved.
In this approximation, the chiller is represented via a simple vapor-compression cycle (see Figure \ref{fig:chiller}). The laws of thermodynamics applied to this cycle under several assumptions, such as steady state, ignoring losses, etc (for the details see \cite{foliaco2020improving}), give:
\begin{align}
&W_{comp} = \frac{-D Q_{ev}^2 - C Q_{ev} - BQ_{ev} + A}{D Q_{ev} + C}, \label{eq:chiller-1}\\
&Q_{ev} - Q_c + W_{comp} + Q_{LT} = 0,\\
&Q_{ev} = C_{ch} (T_{chi} - T_{cho}),\\
&Q_c = C_{cw} (T_{ci} - T_{co}).\label{eq:chiller-4}
\end{align}
Here $W_{comp}$ is the chiller compressor power. $Q_{ev}$ is the cooling load, the heat removed from the building through water flow. $Q_c$ is the removed heat in the condenser. $Q_{LT}$ is a total heat loss. $T_{chi}$ and $T_{cho}$ are the evaporator inlet and outlet temperatures. $T_{ci}$ and $T_{co}$ are the condenser inlet and outlet temperatures. $C_{ch} = \dot{m}_{ch} c_{ch}$ is the thermal capacitance of cooling water ($kW/K$) with $\dot{m}_{ch}$ as a mass flow rate ($kg/s$) and $c_{ch}$ water specific heat capacity. $A, B, C, D$ are constants that encode various other quantities, the exact expressions for them can be found in \cite{foliaco2020improving}. As suggested in \cite{foliaco2020improving}, one can estimate values for these constants upon actual measured values for the chiller.

If we neglect all heat losses $Q_{LT}$, Eqs. (\ref{eq:chiller-1})-(\ref{eq:chiller-4}) become a system of four equations with four unknowns: $Q_{ev}, Q_c, T_{cho}, T_{co}$. Here, the agent provides a recommendation for the chiller leaving temperature $T_{cho}$. Based on this, a low-level PID controller adjusts the compressor speed, hence affecting the compressor power $W_{comp}$ parameter. This adjustment is made to minimize the error between the current sensor reading for $T_{cho}$  with the recommendation provided. Analytically, we are now able to compute the ground truth output temperatures $T_{cho}, T_{co}$ with auxiliary variables $Q_{ev}, Q_c$. Since Eq. (\ref{eq:chiller-1}) is quadratic, in order to compute $Q_{ev}, Q_c$ one needs to select a square root branch. Matching the solution to the read data shows that one must pick the positive branch, this gives the final result:
\begin{align}
&T_{cho} = T_{chi} - \frac{Q_{ev}}{C_{ch}}, \quad T_{co} = T_{ci} - \frac{Q_c}{C_{cw}}, \label{eq:gordon-ng-final-1}\\ 
&Q_{ev} = \frac{-b + R}{2 D}, \quad Q_c = Q_{ev} + W,\\
&b = C + B + D W, \quad R = \sqrt{b^2 - D(CW - A)} \label{eq:gordon-ng-final-3}. 
\end{align}

\begin{figure*}[ht]
    \includegraphics[width=\textwidth]{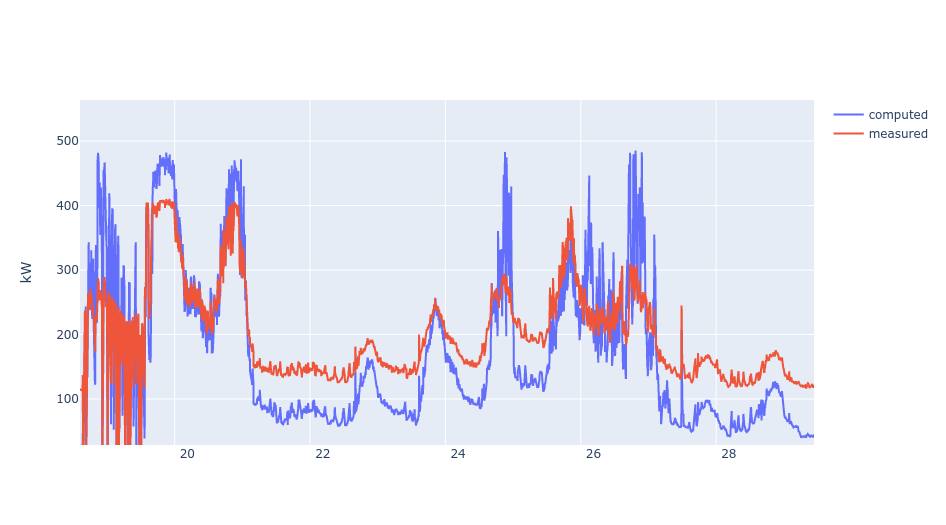}
    \caption{Comparison of the analytical model for a chiller power Eq.(\ref{eq:chiller-1}) and the real observed data (x-axis represents days of a month). While we see a qualitative agreement, the precise match is obtained by tuning the coefficients $A, B, C, D$ (as done in \cite{foliaco2020improving}).}
    \label{fig:chiller-power-match}
\end{figure*}

Figure \ref{fig:chiller-power-match} shows a qualitative match between the chiller power computed via Eq. (\ref{eq:chiller-1}) and the observed data from a real facility. We see that the analytical model reproduces the observed power trends observed in the real facility. A precise match can be obtained by tuning $A, B, C, D$ coefficients (as done in \cite{foliaco2020improving}).

\paragraph{Cooling tower}

\begin{figure*}[h]
\centering
\begin{subfigure}[b]{0.48\textwidth}
     \centering
     \includegraphics[width=\textwidth]{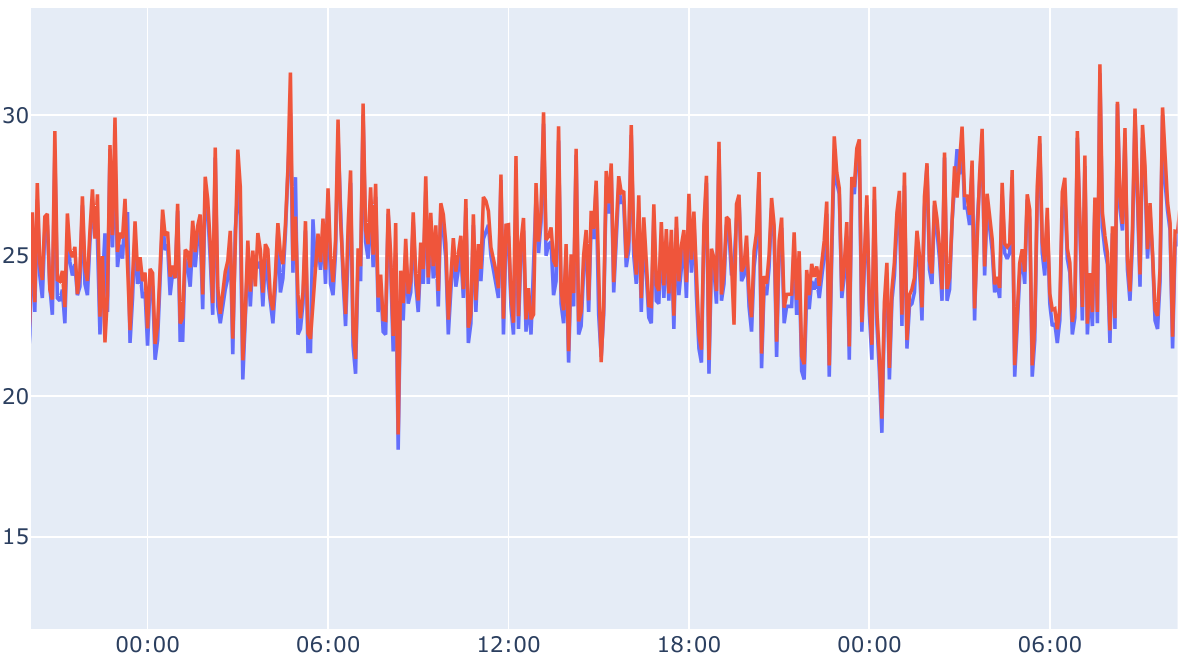}
     \caption{Pump power.}
     \label{fig:pump-power-match}
\end{subfigure}
\hfill
\begin{subfigure}[b]{0.48\textwidth}
     \centering
     \includegraphics[width=\textwidth]{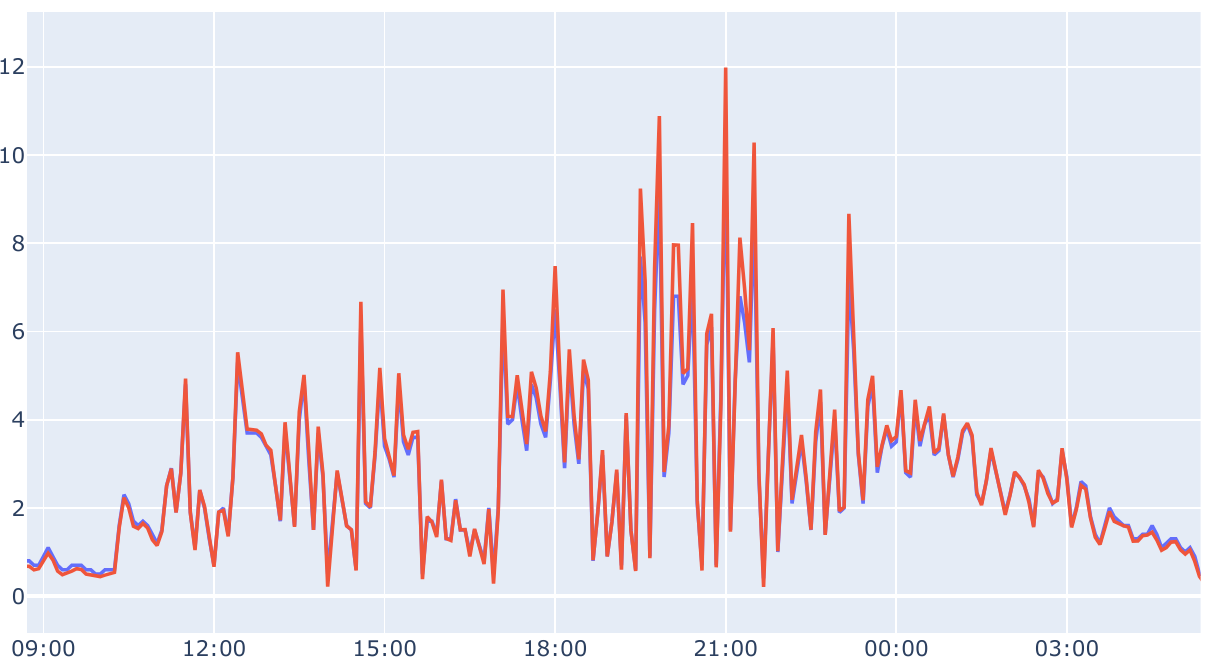}
     \caption{Fan power.}
     \label{fig:fan-power-match}
\end{subfigure}
\caption{Verification of equations for a pump (\ref{eq:cooling-tower-pump}) and a fan (\ref{eq:cooling-tower-fan}) power (x-axis represents times of day). Red lines represent computed values and blue lines - measured.}
\label{fig:fan-pump-power-match}
\end{figure*}

Using the laws of thermodynamics, it was shown in \cite{cao2018optimal} that the temperature of the water leaving the cooling tower can be computed as:
\begin{equation}
\label{eq:cooling-tower-temp}
T_{co} = T_{ci} - (T_{ci} - T_{wb})(1 - \mathrm{exp}[c_8 (P_{pump})^{c_9} (P_{fan})^{c_{10}}]),
\end{equation}
where $T_{ci}$ is the temperature of the water entering the cooling tower, $T_{wb}$ is the wet-bulb temperature, $P_{pump}$ is the condenser water pump frequency (speed), $P_{fan}$ is the cooling tower fan frequency (speed). As the valve position on cooling water pipe and the resistance characteristics of the cooling water pipe unchanged, the condenser water pump flow is proportional to the water pump frequency.
Similarly, the cooling tower air flow and fan power are also consistent with the law. The condenser water flow rate $m_c$, the condenser water pump power consumption $W_{pump}$, the cooling tower air flow $m_a$ and the cooling tower fan power consumption $W_{fan}$ are calculated \cite{cao2018optimal} by:
\begin{align}
&m_c = c_{11} P_{pump}, \quad W_{pump} = c_{12} (P_{pump})^3,\label{eq:cooling-tower-pump}\\
&m_a = c_{13} P_{fan}, \quad W_{fan} = c_{14} (P_{fan})^3.\label{eq:cooling-tower-fan}
\end{align}
The coefficients $c_i$ are set by fitting the observed data (this can be done via a regression as described in\cite{foliaco2020improving}). Figure \ref{fig:fan-pump-power-match} illustrates the  matching dynamics between equations for the fan/pump power and the observed data.

\paragraph{Multiple cooling towers}

\begin{figure*}[h!]
\centering
\begin{subfigure}[b]{\textwidth}
     \centering
     \includegraphics[width=\textwidth]{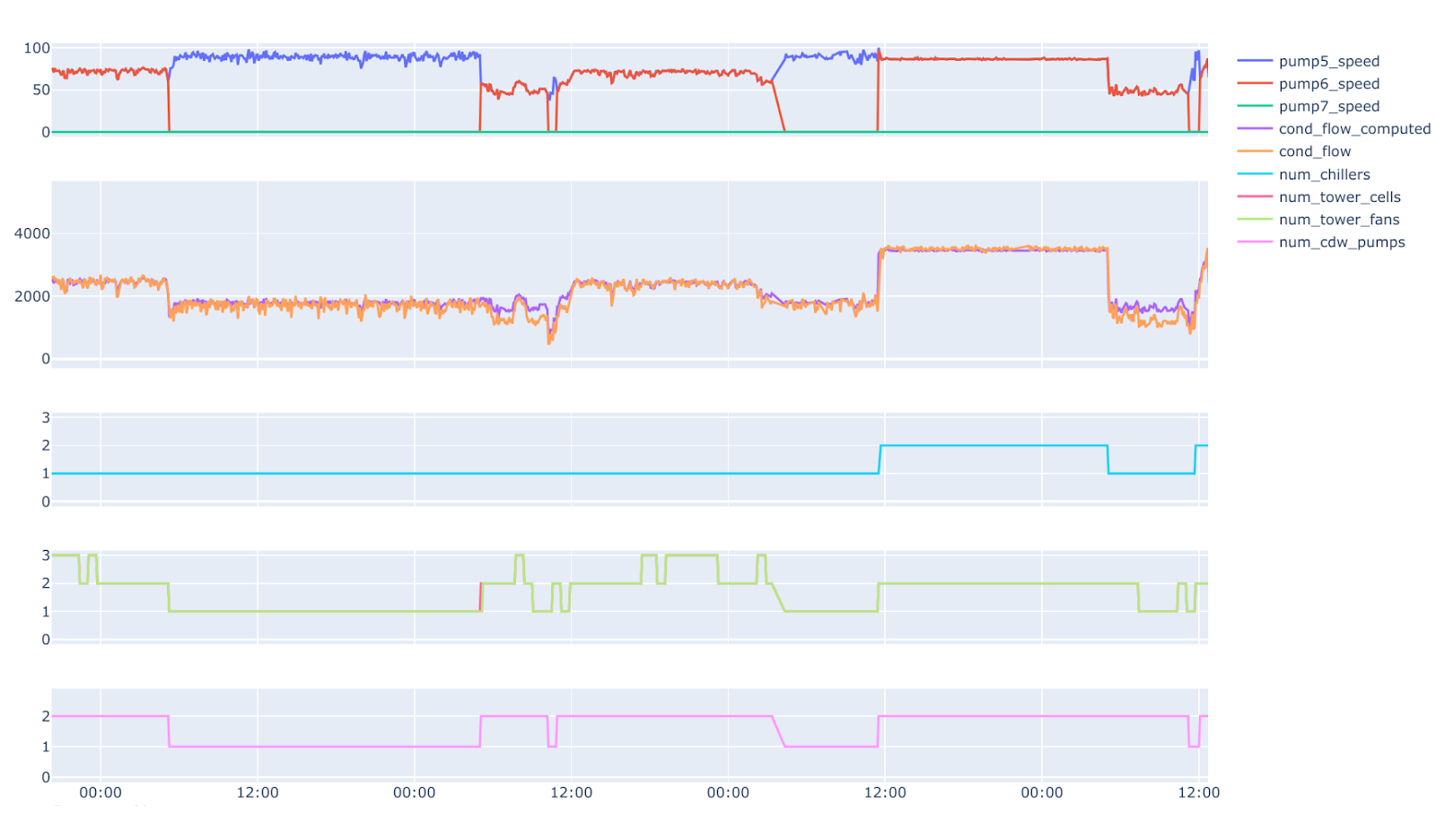}
     \caption{Short timescale (x-axis represents times of day).}
     \label{fig:condenser-water-flow-match-short-time-scale}
\end{subfigure}
\vfill
\begin{subfigure}[b]{\textwidth}
     \centering
     \includegraphics[width=\textwidth]{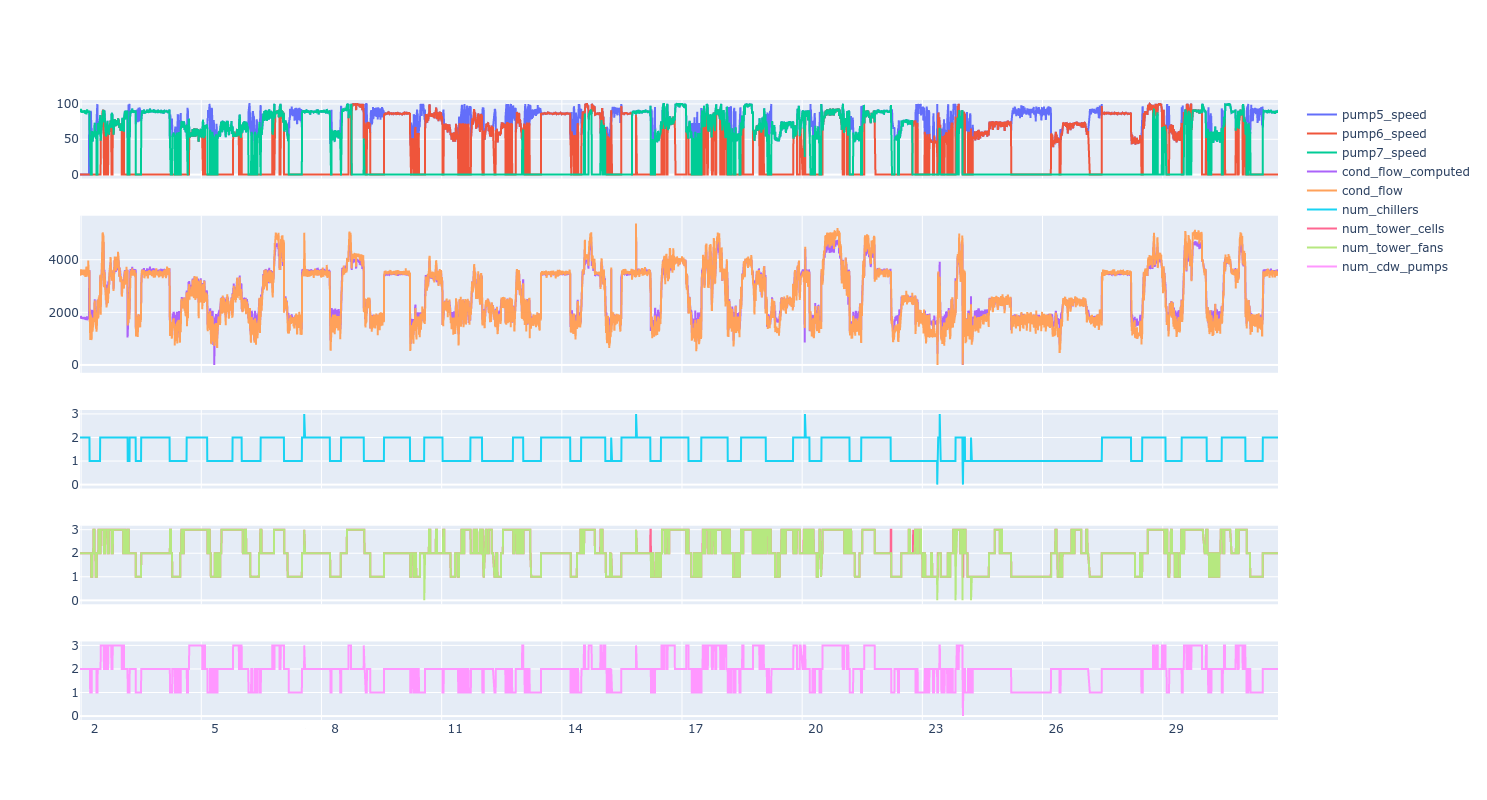}
     \caption{Long timescale (x-axis represents days).}
     \label{fig:condenser-water-flow-match-long-time-scale}
\end{subfigure}
\caption{Verification of Eq.(\ref{eq:cond-water-flowrate-multiple-cooling-towers}) for the condenser water flow rate under different circumstances. The second row on each figure shows a good agreement between computed vs observed flow rates for both (a) short and (b) long time scales. Figures in other rows show that the good agreement is observed for various situations including different number of chillers, fans and pumps. Here we have three pumps labeled as pump5, pump6, pump7.}
\label{fig:condenser-water-flow-match}
\end{figure*}

Real-world cooling facilities usually have multiple cooling towers. In order to describe a real facility we need to extend Eqs.(\ref{eq:cooling-tower-pump}), (\ref{eq:cooling-tower-fan}) to multiple cooling towers. To do so, we observe that the pumps create a force to push the water, and that there are several obstructions that influence the flow. When the number of chillers and pumps are the same, we get a perfect linear relationship; when there are more pumps, the chillers' flow is weaker due to resistance. To account for this, we can introduce another term in the first equation in Eq. (\ref{eq:cooling-tower-pump}), which is proportional to the difference between the number of chillers and pumps: 
\begin{equation}
\label{eq:cond-water-flowrate-multiple-cooling-towers}
m_c = \sum_i^N P_{pump}^{(i)} (a_1 - a_2(N_{pumps} - N_{chillers})).
\end{equation}
Note that the number of fans might have a small effect on this as well. It turns out that this effect is negligible, as the suggested equation holds in the observed data (see Figure \ref{fig:condenser-water-flow-match}).

A further analysis shows that Eq.(\ref{eq:cooling-tower-temp}) can be easily extended for multiple pumps/fans via a simple replacement $P
\to\sum_i P^{(i)}$:
\begin{equation}
\label{eq:cooling-tower-temp-multiple-cooling-towers}
T_{co} = T_{ci} - (T_{ci} - T_{wb})\left(1 - \mathrm{exp}\left[c_8 \left(\sum_i P_{pump}^{(i)}\right)^{c_9} \left(\sum_{j} P_{fan}^{(j)}\right)^{c_{10}}\right]\right).
\end{equation}

Now, we apply the above equations to the chiller plant depicted at Figure \ref{fig:chilled-water-plant-overview-detailed}. We need to use inverse variations on the above equations, such that the inputs are the setpoints (plus other measured variables), and the outputs are the condenser water pump power ($W_{pump}$) and the fan power ($W_{fan}$). In our chiller plant, all pumps (and fans) are run with the same frequencies, so re-arranging (\ref{eq:cond-water-flowrate-multiple-cooling-towers}), (\ref{eq:cooling-tower-fan}), (\ref{eq:cooling-tower-temp-multiple-cooling-towers}) gives:
\begin{align}
&P_{pump}^{(i)} = P_{pump} = \frac{m_c}{c_{11}N_{pumps}(a_1 - a_2(N_{pumps} - N_{chillers}))},\\
&W_{pump}^{(i)} = c_{12} (P_{pump})^3, \quad i\in [1,\cdots, N_{pumps}].\\
&P_{fan}^{(j)} = P_{fan} = \frac{1}{N_{fans}} \left(\frac{1}{c_8 (N_{pumps}P_{pump})^{c_9}} \log\left[ 1 + \frac{T_{co} - T_{ci}}{T_{ci} - T_{wb}} \right] \right)^{1/c_{10}},\\
&W_{fan}^{(j)} = c_{14} (P_{fan})^3, \quad j\in [1,\cdots, N_{fans}].
\end{align}

\paragraph{Integrating Analytical Solutions into Multi-physics Software}

As described in the beginning of this section, we use a commercially available multi-physics software for modeling the fluid dynamics. This offers the flexibility to consider a wide variety of different physical processes/applications that can be solved as transient or steady state problems. In our case this software allows to define source and sinks for the physical parameters such as temperature, fluid velocity and pressure. Given the transient simulation, we extract the measured parameter values at every simulation step. Hence, we leverage the simulation state variables which serve as a subset of input parameters for the analytical equations described above.

For example, in case of the Gordon-Ng chiller model, we wish to compute the temperature gradient at both the evaporator and the condenser side (see Figure \ref{fig:chiller}). We consider the measured chiller inlet temperatures and mass flow rate at each of these ends. These are measured via "sensor probes" attached at the inlet of the chiller. At every simulation step, these readings are fed into the system of equations (\ref{eq:gordon-ng-final-1})-(\ref{eq:gordon-ng-final-3}) and the output temperature values are computed. We assume a negligible fluid flow, pressure drop across the chiller cross section (relative to that across the facility). Next, we define a fluid flow source, which acts as the chiller outlet. The incoming water temperature is also provided by the output of the Gordon-Ng model (\ref{eq:gordon-ng-final-1})-(\ref{eq:gordon-ng-final-3}). We constrain this outlet incoming differential pressure \footnote{The differential pressure is defined as the difference in pressure between two points in the water pipes. In the case of the chilled water system, the differential pressure at any point is measured relative to the pressure at the chilled water pump inlet.} and mass flow rate to that of the inlet measured values. This process is repeated for all chillers.

The same technique is used for the cooling tower and pump bank analytical models. The former only computes a temperature gradient similar to the chiller, and the pump bank computes the differential pressure.

\subsection{Simulation Fidelity}
\label{sec:simulation-fidelity}

In the previous section, we summarized analytical solutions used as a part of our simulation and demonstrated their validity against real-world data. In this section we provide some additional fidelity checks for the entire simulation, which combines the analytical solutions with simulation software libraries as described above.

\begin{figure}[h!]
\centering
\begin{subfigure}[b]{0.48\textwidth}
     \centering
     \includegraphics[width=\textwidth]{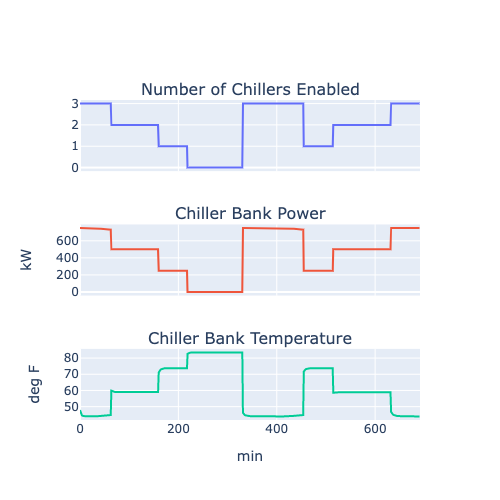}
     \caption{Changing the number of chillers.}
     \label{fig:simulation-num-chiller-change}
\end{subfigure}
\begin{subfigure}[b]{0.48\textwidth}
     \centering
     \includegraphics[width=\textwidth]{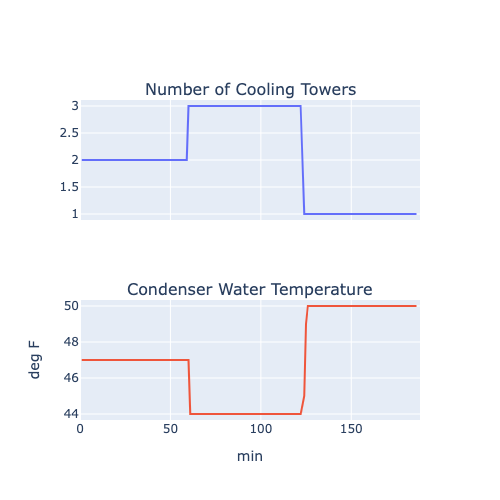}
     \caption{Changing the number of cooling towers.}
     \label{fig:simulation-num-cooling-towers-change}
\end{subfigure}
\vfill
\begin{subfigure}[b]{0.48\textwidth}
     \centering
     \includegraphics[width=\textwidth]{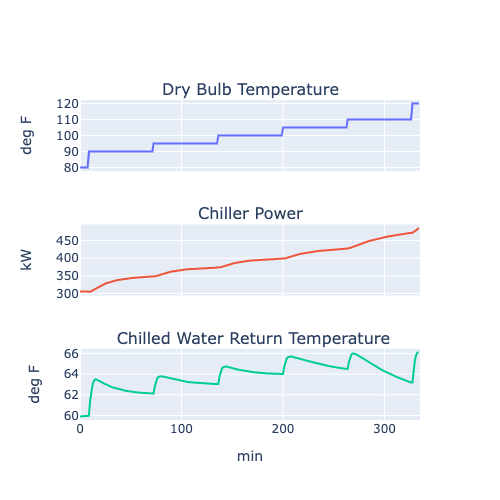}
     \caption{Increasing the dry bulb temperature.}
     \label{fig:simulation-dbt-change}
\end{subfigure}
    
\caption{High level simulation fidelity check. Increasing the number of chillers (a) or (b) number of cooling towers decreases the overall temperature in the system. Similarly, in (c) an increase in the external temperature, increases the chiller power consumption.}
\label{fig:simulation-simple-fidelity-checks}
\end{figure}

We start with simple demonstrations in Figure \ref{fig:simulation-simple-fidelity-checks}. First, keeping all setpoints fixed, we vary the number of chillers. As expected, turning on more chillers decreases the system's temperature and turning off chillers have the opposite effect (see Figure \ref{fig:simulation-num-chiller-change}). Second, we vary the number of cooling towers. Again, as expected, turning on more cooling towers decreases the system's temperature and other way around - turning off cooling towers increases the temperature (see Figure \ref{fig:simulation-num-cooling-towers-change}). As a slightly different example, we fix the chiller leaving temperature setpoint and increase the dry bulb temperature, which represents the external weather conditions. As expected, we observe (Figure \ref{fig:simulation-dbt-change}) an increase in the compressor power required to maintain the chiller temperature. After leaving the air handling unit (AHU), the water gets progressively warmer due to the greater heat generated from the building (which is modeled as the increase in the dry bulb temperature). We also see the intermediate transitions in the temperature as the PID controllers try to control the chiller power. After a rapid transition due to the increase in dry bulb temperature, the PIDs react by reducing the power.

\begin{figure}[h!]
\begin{subfigure}[b]{\textwidth}
    \includegraphics[width=\textwidth]{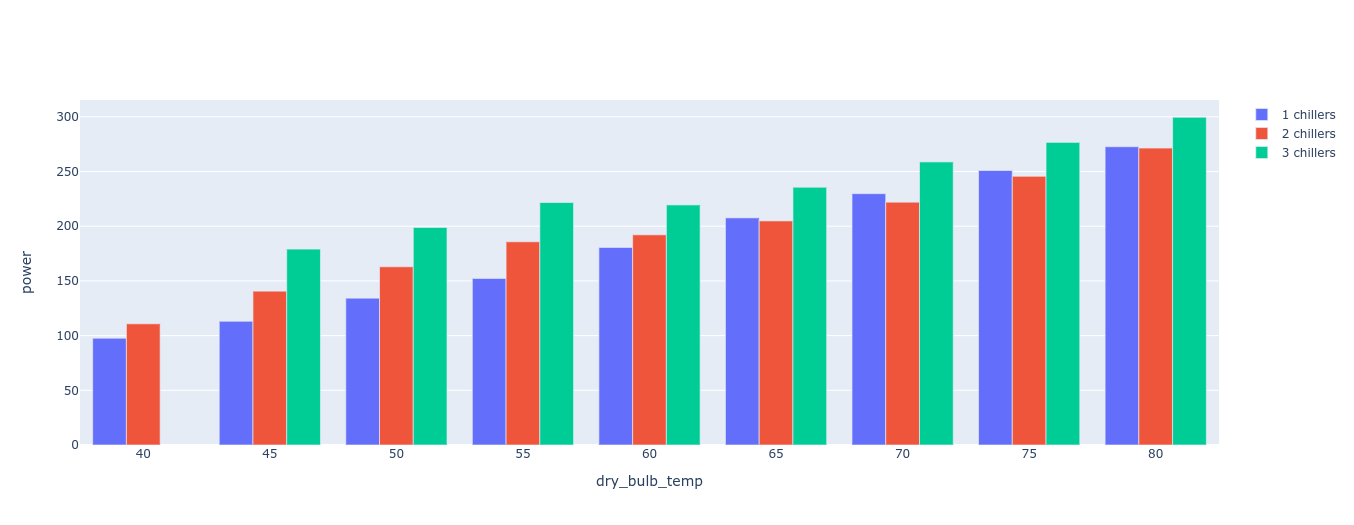}
    \caption{Simulated data.}
    \label{fig:simulator-power-vs-dry-bulb-chillers}
\end{subfigure}
\vfill
\begin{subfigure}[b]{\textwidth}
    \includegraphics[width=\textwidth]{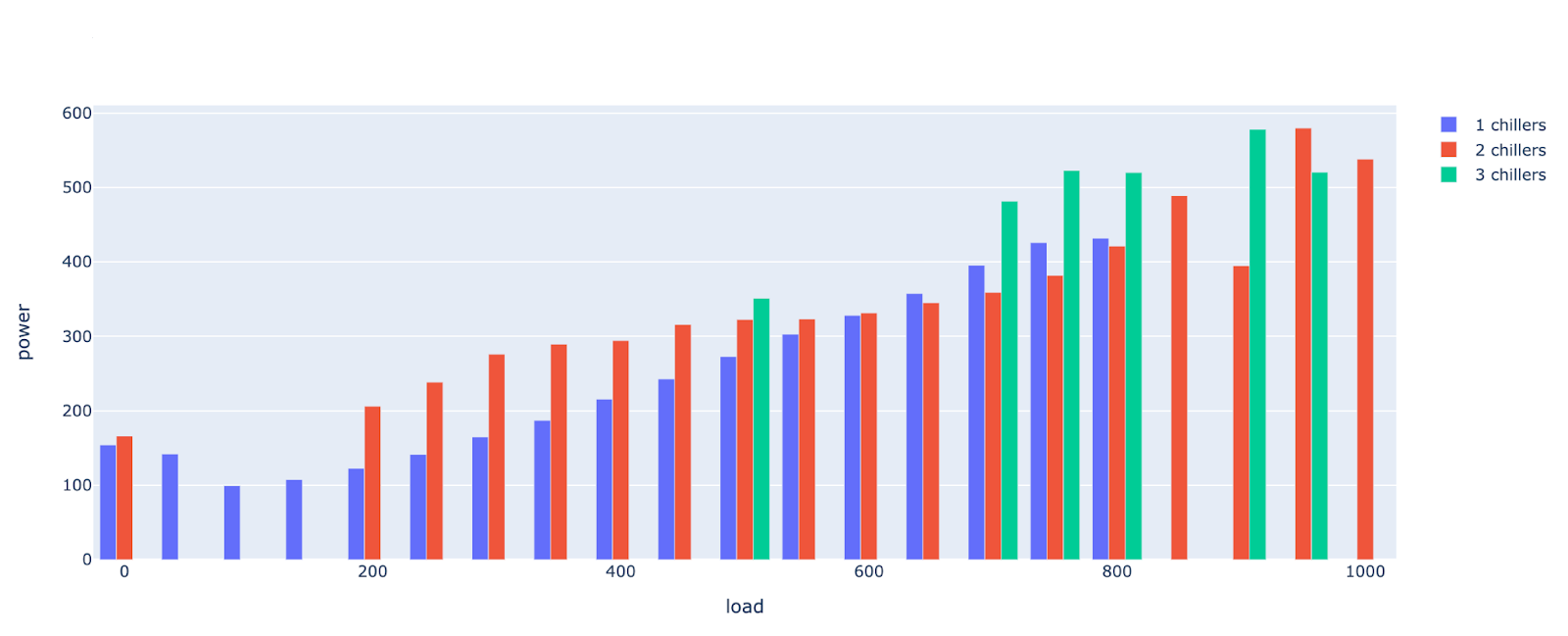}
    \caption{Real data.}
    \label{fig:real-data-power-vs-load}
\end{subfigure}
\caption{Total power consumed by all chillers as a function of dry-bulb temperature (load). We see that for low dry bulb temperatures (load) it is always more energy-efficient to use one chiller. Two chillers become more efficient for high values of the dry-bulb temperature (load).}
\label{fig:simulator-check-emergent-chiller-power}
\end{figure}

Next, we demonstrate that our simulation exhibits a more complex "emergent" behavior. Since chillers are usually the most energy-intensive components of any cooling facility, it is important to efficiently control how many chillers are used at any given time. The domain experts know that depending on external conditions, there are regimes when it is more energy-efficient to use two chillers even though a single chiller can achieve the desired temperature.
To show that our model encodes this information, we plot the total power used by all chillers as a function of dry bulb temperature. For each value of the dry bulb temperature we show the total power used by different number of chillers (see Figure \ref{fig:simulator-power-vs-dry-bulb-chillers}). This needs to be compared to the real world data. To do so, we display the total power used by all chillers as a function of load (in our simulated environment dry-bulb temperature is a proxy for the load). For each value of the load we show the total power used by different number of chillers. For low values, one chiller always consumes the lowest amount of energy. However, for higher values two chillers are more energy efficient. We observe the quantitative match between the simulated and observed transitions from real world data.

\section{Reinforcement Learning Applications}

To demonstrate how our industrial cooling system model (Section \ref{sec:simulation}) can be used for reinforcement learning applications, in this section we introduce the Industrial Task Suite, define several tasks for the model, and use these tasks to evaluate the performance of different RL algorithms.

\subsection{Industrial Task Suite}
\label{sec:its}

\begin{figure*}[h]
    \includegraphics[width=\textwidth]{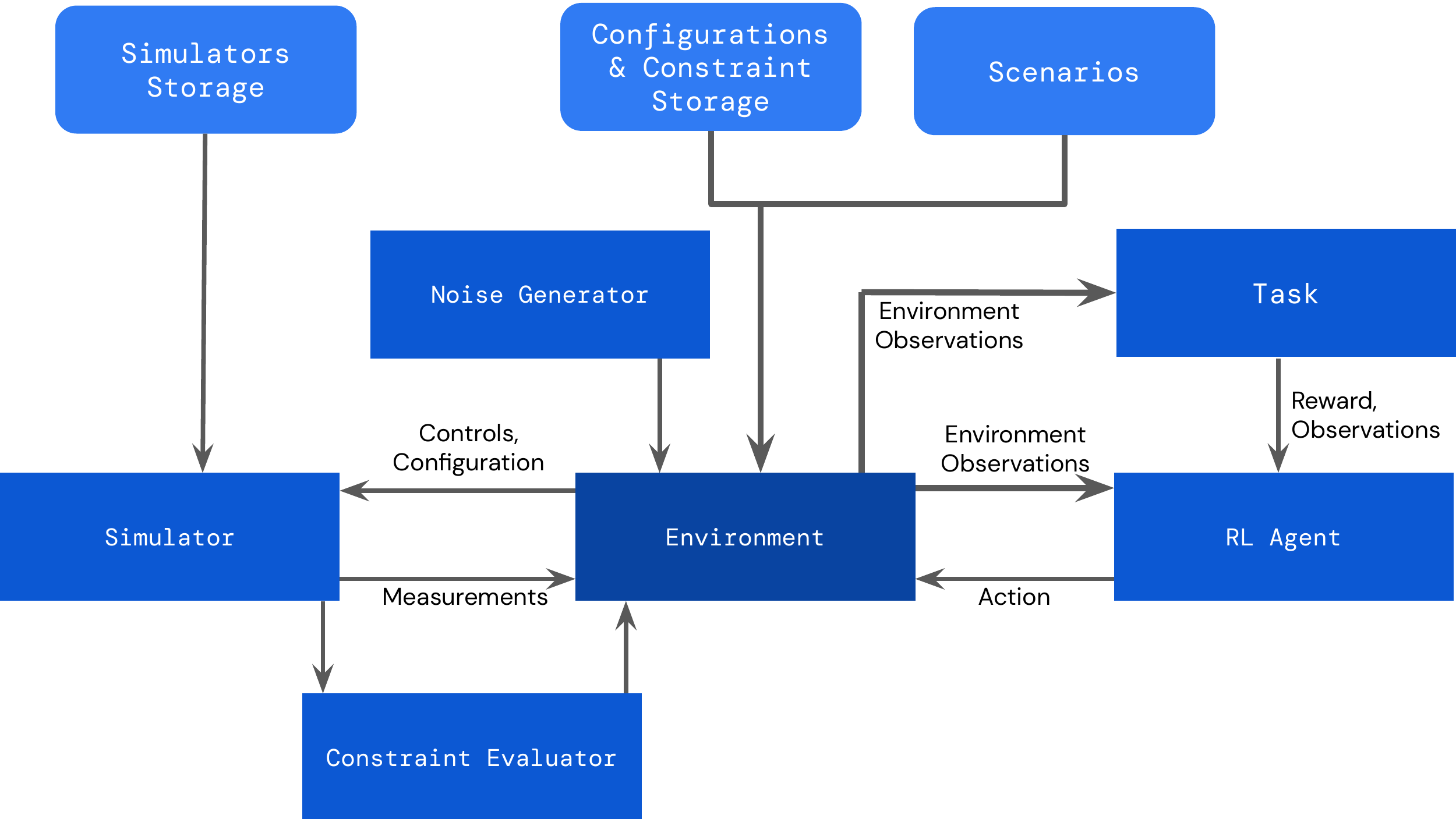}
    \caption{The Industrial Task Suite design diagram. The \texttt{Environment} class orchestrates the main components: simulator, noise generator, task, constraint evaluator and agent. It takes a configuration and uses it to initialize a simulation loaded from the simulator database. It then starts relaying the traffic between the agent and the simulator.}
    \label{fig:its-design}
\end{figure*}

We begin with the Industrial Task Suite, a flexible framework for defining industrial control problems within a reinforcement learning context. 
It allows to create environments and tasks to train and evaluate artificial controllers safely and efficiently, subject to various real world conditions, such as sensor imperfections and white noise. ITS is in-line with the design of the DeepMind Control Suite \cite{tassa2018deepmind} and Real-World Reinforcement Learning (RWRL) Challenge Framework \cite{dulac2019challenges}, while providing more granular configurability of environments.
Balancing ease-of-use and customization, the task suite only requires a simulation (model) and a task to define a RL problem (Figure ~\ref{fig:its-design}), and yet offers a highly parameterized facility configuration, which provides industrial equipment components with customizable capacities, amounts and connection topology. The task suite also supports specific scenarios such as sensor noise/drift, weather variations, and varying initial conditions to mimic specific real-world operating regimes. A diversity of such "frozen" scenarios could serve as a test bed for evaluating performance and benchmarking against existing baselines.

The core \texttt{Environment} class encapsulates and orchestrates the following components: \textit{simulator}, \textit{noise generator}, \textit{task}, \textit{constraint evaluator} and \textit{agent} (see Figure \ref{fig:its-design}). It takes a configuration from the database and uses it to initialize a simulation. Then \texttt{Environment} applies one or several implemented scenarios and defines the constraints. It uses a pre-defined \textit{task} to compute rewards and expand the observations space with additional information. It then starts relaying the traffic between the agent and the simulation. While doing so, it injects noise into the exchanged data, evaluates the constraints and calculates the rewards.

\texttt{Environment} converts the inbound actions sent by an agent into simulation control signals, and subsequently convert the outbound simulation measurements into observations. Given a broad range of possible control/measurement specs, the converters pad the values to a defined maximum range and append masks to the observation to inform the agent about the differences in the configuration. In order to provide access to all of the available information, the facility configuration and the spec definition is also added to the observation dictionary. Should that information be obsolete for a particular agent architecture, it can be filtered out in an environment wrapper.

\textit{Noise} is added to the control signal and to the observation respectively. The noise simulates sensor/pipeline imperfections, and in effect diversifies the distribution of simulated states (control noise) and observations (observation noise).
At each episode step, in addition to sending the control signal, \texttt{Environment} also changes the values of the selected configuration parameters according to pre-defined \textit{scenarios}. This allows simulating the changing environment conditions and other environmental instabilities.

A more detailed overview of ITS's main components along with some implementation details are given in Appendix \ref{sec:its-components}.

\subsection{Tasks}
\label{sec:HVAC-tasks}

The goal of an AI-powered system that operates a cooling facility is to minimize the energy consumption while meeting a set of constraints. For industrial cooling systems these constraints are defined to ensure that buildings are comfortable for the people inside and to prevent the equipment breakage. To efficiently train RL agents within the ITS framework, we consider a curriculum learning based approach \cite{bengio2009curriculum} (for a recent review see \cite{narvekar2020curriculum}). Namely, we define various tasks categorized by their perceived complexity and teach agents to solve them sequentially. Eventually, we will have an agent that knows how to solve all the tasks and operate the cooling facility.

We start with action (Table \ref{table:action-spec}) and observation (Table \ref{table:obs-spec}) space defined for our simulator (Section \ref{sec:simulation}). All values are selected based on a typical industrial cooling facility configuration.

\begin{table}[]
\centering
\resizebox{\textwidth}{!}{%
\begin{tabular}{|l|l|l|l|l|l|}
\hline
Action                                 & Unit  & Type       & Default & Minimum & Maximum \\
\hline
\# Chillers Enabled                    & -     & Integer    & 1       & 0       & 3       \\
\# Chilled Water Pumps                 & -     & Integer    & 1       & 1       & 3       \\
\# Condenser Water Pumps               & -     & Integer    & 1       & 1       & 3       \\
Chiller Leaving Temperature            & deg F & Continuous & 48      & 40      & 75      \\
Cooling Tower Return Water Temperature & deg F & Continuous & 55      & 32      & 90      \\
Condenser Water Pump Flow rate         & kg/s  & Continuous & 50      & 10      & 200     \\
Distributional Differential Pressure   & psi   & Continuous & 15      & 0.1     & 50      \\
\# Free cooling Heat Exchangers        & -     & Integer    & 1       & 1       & 3   \\ \hline 

\end{tabular}%
}
\caption{Action space used in our experiments.}
\label{table:action-spec}
\end{table}

\begin{table}[H]
\centering
\begin{tabular}{|l|l|l|}
\hline
Observation                                 & Type       & Unit  \\
\hline
Building load                               & Continuous & kW    \\
Dry bulb temperature                        & Continuous & deg F \\
Wet bulb temperature                        & Continuous & deg F \\
Relative humidity                           & Continuous & -     \\
Condenser leaving temperature (per chiller) & Continuous & deg F \\
Chilled water flow rate (per chiller)       & Continuous & kg/s  \\
Compressor power (per chiller)              & Continuous & kW    \\
Cooling tower fan power                     & Continuous & kW    \\
Condenser water pump bank power             & Continuous & kW    \\
Chilled water pump bank power               & Continuous & kW    \\
Chiller bank leaving temperature            & Continuous & deg F \\
Chilled water supply temperature            & Continuous & deg F \\
Chilled water return temperature            & Continuous & deg F \\
\# of chillers enabled                      & Integer    & -    \\
\hline
\end{tabular}%
\caption{Observation space used in our experiments.}
\label{table:obs-spec}
\end{table}

To mimic the industrial setting, the constraints are split into soft and hard. A hard constraint violation terminates the episode (the agent gets kicked out of the environment), which incentivizes the agent to respect the constraint. Soft constraints can be added to the reward function via additional terms (for more see Section \ref{sec:reward-function}). Additionally, any constraint violation is communicated to the agent as an observation. For more about constraints see Section \ref{sec:constraints}.

Next, we list several tasks sorted by difficulty and provide (if known) corresponding baseline and the optimal policies. 

\paragraph{Easy}\hfill \break

\noindent\textbf{Unconstrained number of chillers control} \textit{Controls}: number of chillers (1D integer). \textit{Baseline policy}: use 0 chillers. \textit{Optimal policy}: use 0 chillers. This is a very simple task, where we do not impose any constraints and not worry about any temperatures in the system. Ultimately, controlling the number of chillers is the most important aspect of a cooling facility control, because chillers are the most power-consuming components.

\noindent\textbf{Constrained number of chillers control} \textit{Controls}: number of chillers (1D integer). \textit{Hard constraint}: use only a subset of available chillers, for example, if there are 3 available chillers, the agent is allowed to use only 1 or 2 chillers. \textit{Baseline policy}: use 1 chiller. \textit{Optimal policy}: use 1 chiller. This is a slightly more complicated chiller control task, where we still do not worry about the temperatures, but impose the constraint on the number of chillers. Via this task we explore whether the agent can learn to satisfy hard constraints.

\noindent\textbf{Chiller temperature control} \textit{Controls}: chiller temperature (1D float). \textit{Baseline policy}: use the highest possible chiller temperature. \textit{Optimal policy}: use the highest possible chiller temperature. In a typical cooling facility, along with the number of chillers, one can control individual chiller temperatures. With this task we investigate whether the agent can learn to effectively control them.

\paragraph{Medium}\hfill \break

\noindent\textbf{Constrained number of chillers control with an additional constraint on the chilled water supply temperature} \textit{Controls}: number of chillers (1D integer). \textit{Hard constraint}: use only a subset of available chillers and constraint on the range of chilled water supply temperature. \textit{Scenario}: Randomized dry-bulb temperature. \textit{Baseline policy}: use 1 chiller. \textit{Optimal policy}: depending on the external conditions use 1 or 2 chillers. This task is closer to real facility requirements; the agent controls the number of chillers while keeping the chilled water supply temperature within a given range, which directly translates to the overall temperature in a building and its occupants' comfort.

\noindent\textbf{Number of chillers and condenser temperature control with an additional constraint on the chilled water supply temperature} \textit{Controls}: number of chillers (1D integer) and condenser temperature (1D float). \textit{Hard constraint}: constraint on the range of chilled water supply temperature. In this task, in addition to the number of chiller control, the agent controls the condenser temperature while keeping the chilled water supply temperature within a given range.

\paragraph{Hard}\hfill \break

\noindent\textbf{Full cooling system control} \textit{Controls}: All actions defined in Table \ref{table:action-spec}. \textit{Hard constraint}: constraint on the range of chilled water supply temperature. In this task the agent controls all available cooling facility setpoints while keeping the chilled water supply temperature within a given range.

\paragraph{Other tasks}\hfill \break

For a fixed simulator the new tasks can be constructed by using one or a combination of several possible ways:
\begin{itemize}
    \item Selecting a task objective (for example, minimize the power consumption).
    \item Adding a scenario (for example, changing weather).
    \item Selecting a subset of controls (for example, only control the number of chillers).
    \item Adding noise to controls.
    \item Adding noise to measurements.
    \item Imposing constraints.
\end{itemize}

\subsection{Reward Function}
\label{sec:reward-function}

As discussed in Section \ref{sec:HVAC-tasks}, we split constraints into soft and hard, where a hard constraint violation terminates the episode, and soft constraints are added to a reward function via additional terms. Moreover, any constraint violation is communicated to the agent as an observation. The base reward function (without soft constrains) should satisfy the following requirements:
\begin{enumerate}
    \item Inverse relationship between power and reward (more power - less reward).
    \item Maximum reward = 1 (when the power is 0, which is the best case scenario, the reward is 1).
    \item Reward function is sensitive to a change in the relevant range of power (this depends on the cooling system configuration).
\end{enumerate}

\begin{figure*}[h]
    \centering
    \includegraphics[width=0.8\textwidth]{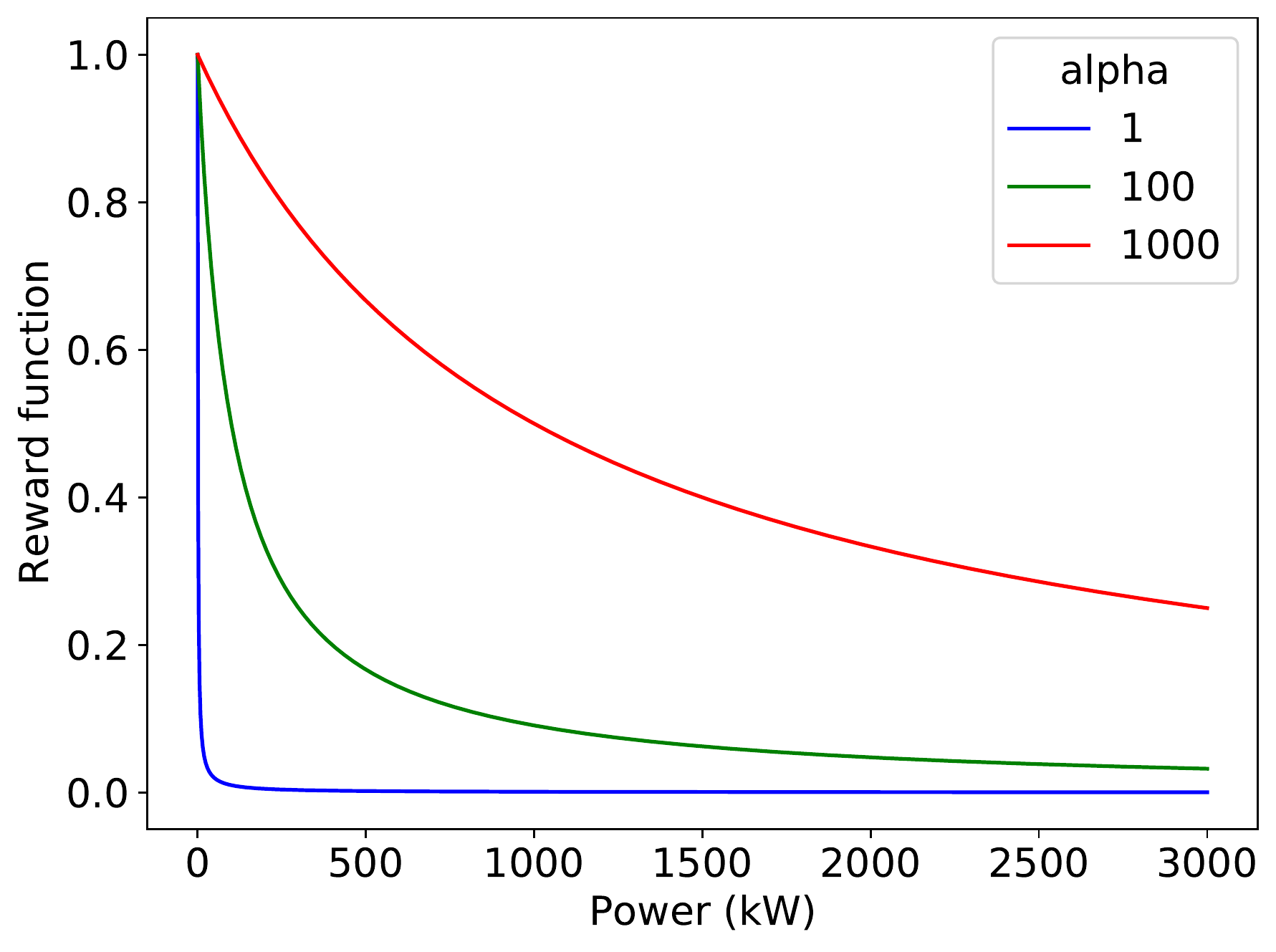}
    \caption{Base reward function for power optimization tasks - Eq. (\ref{eq:hvac-reward-function}). The normalization constant, $\alpha$, allows to make the reward function sensitive to the power change in the relevant range of values. In this example, when $\alpha = 1$ the reward function has nearly the same value for the entire 10-3000 kWatt range. $\alpha = 1000$ provides a good gradual reward change from 10 to 3000 kWatts.}
    \label{fig:hvac-reward-function}
\end{figure*}

The following simple function satisfies all the requirements:
\begin{equation}\label{eq:hvac-reward-function}
r = \frac{1}{W/\alpha + 1}.
\end{equation}
Here $W$ is the power consumed by the system and $\alpha$ is a normalization constant. $1$ in the denominator ensures that $r(0) = 1$ and the normalization constant ($\alpha$) makes the function sensitive to the relevant range of power. For example, for industrial chillers $\alpha \sim 1000$ in the SI units (kWatt).

\subsection{Experiments and Results}
\label{sec:experiments-and-results}

We now investigate whether the common deep RL algorithms can be used for the industrial cooling facility control by benchmarking these algorithms on some of the tasks defined in Section \ref{sec:HVAC-tasks}. We consider DMPO \cite{acme-DMPO-agent, abdolmaleki2018maximum}, DDPG \cite{lillicrap2015continuous}, D4PG \cite{barth2018distributed}, and use their ACME implementations \cite{hoffman2020acme}. For some of the tasks we know the optimal policy and will see if the RL agents are able to learn it.

\begin{figure*}[h]
\centering
\begin{subfigure}[b]{\textwidth}
     \centering
     \includegraphics[width=\textwidth]{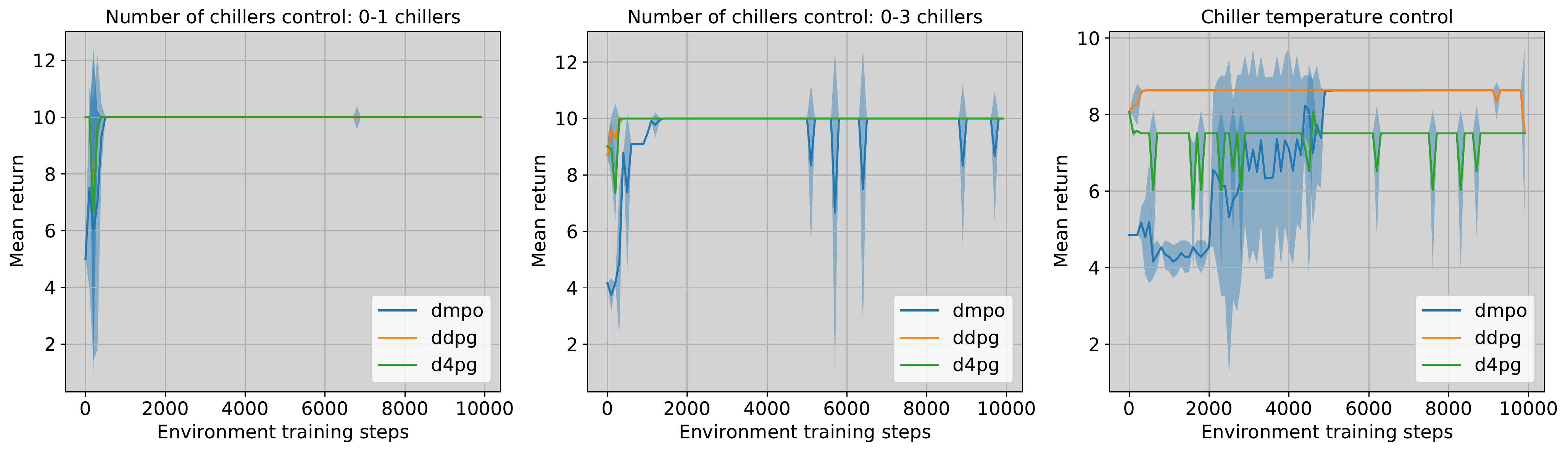}
     \caption{Easy tasks.}
     \label{fig:hvac-results-easy-tasks}
\end{subfigure}
\vfill
\begin{subfigure}[b]{\textwidth}
     \centering
     \includegraphics[width=\textwidth]{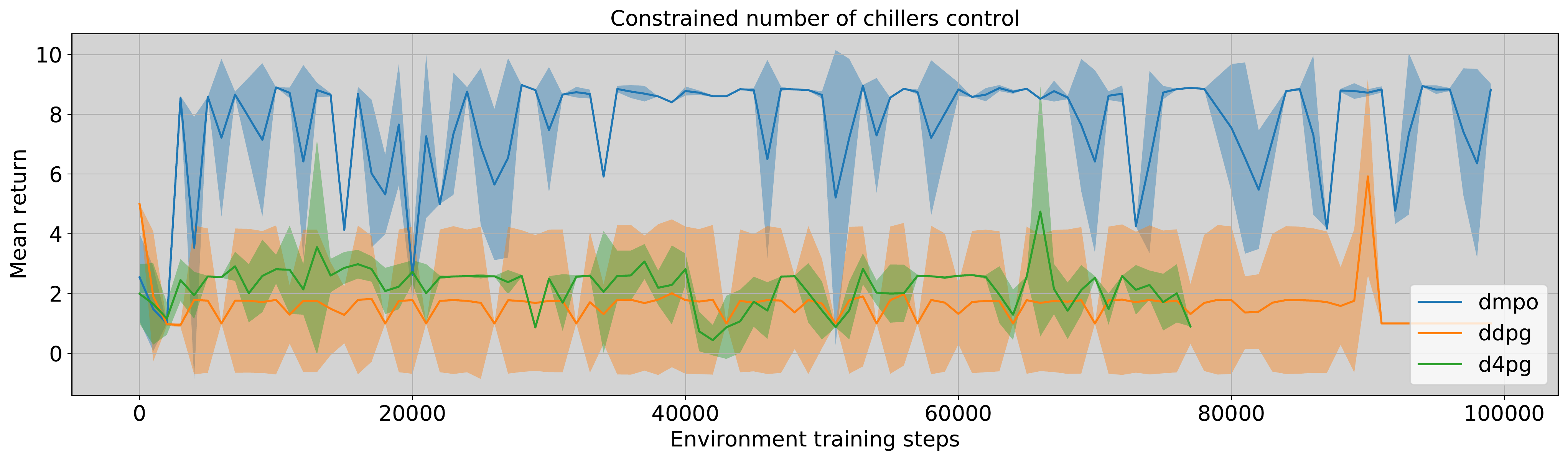}
     \caption{Medium task.}
     \label{fig:hvac-results-medium-tasks}
\end{subfigure}
\caption{Comparison of DMPO, DDPG and D4PG on the tasks defined for our simulated cooling system environment. We consider three easy and one medium tasks defined in Section \ref{sec:HVAC-tasks}. Easy tasks include: unconstrained chiller control (a binary 0-1 chillers and non-binary 0-3 chillers) and a chiller temperature control. For the medium task we use the constrained number of chiller with an additional constraint on the chilled water temperature. Note that for each task we run an experiment with three seeds and report the mean return and uncertainty (shaded area). Interestingly, DMPO is the only agent that learned the optimal policy for all the tasks (for optimal policies and max returns see Section \ref{sec:HVAC-tasks}).}
\label{fig:hvac-experiment-results}
\end{figure*}

For all agents we employ the continuous action space normalized to [-1, 1]. If controls are integer-valued we quantize the agent's continuous actions\footnote{This is done for simplicity since the agents we consider are designed for continuous actions.}. Each agent was run 3 times per task using different seeds. For all tasks we use episodes of length 10. Figure \ref{fig:hvac-experiment-results} shows performance on individual tasks, where the lines denote the mean return and the shaded regions denote standard deviation across seeds. For easy tasks we use 10 actors per agent and for the medium task - 100. We used the same hyperparameters across all tasks (i.e. so that nothing is tuned per-task). Due to a slow simulator, we had to use mini-batches of size 4 across all tasks, and frequently update the actor networks - every 10 actor steps.

The summary of results is presented on Figure \ref{fig:hvac-experiment-results}. For our experiments we consider tasks for controlling the number of chillers and the chiller temperature. As motivated in Section \ref{sec:HVAC-tasks}, these are the most important setpoints for an effective cooling facility control. For the number of chillers control, we start with a very simple binary configuration: the agent can select 0 or 1 chiller. All agents learn this trivial task very fast (in $\sim$100 training steps). Next, we use the configuration with 4 states: the agent can choose between 0, 1, 2, and 3 chillers. And again all agents learned this task easily, although DMPO took a bit longer while also experiencing a few spikes in the mean return across different seeds. For the chiller temperature control, DDPG converges to the optimal policy very fast, while D4PG converges to a sub-optimal policy across all seeds. DMPO does learn the optimal policy, but again it converges slower than DDPG. Finally, we consider a medium difficulty task: the agent controls the number of chillers (0, 1, 2, 3 chillers are available), but if it selects 0 or 3 chillers the episode terminates; there is also additional constraint on the chiller water supply temperature, which directly maps to the overall building temperature. In this case DMPO is the only agent to learn the optimal policy. Overall, we see that DMPO performs the best, as it learns the optimal policies for all four tasks. Next, we provide more details on the architectures for each agent.

\paragraph{DMPO}

Both policy and critic networks are LayerNorm MLPs - they are MLPs with ELU non-linearities preceded by a linear layer followed by a LayerNorm and $\tanh$ activation: $Linear \to LayerNorm \to tanh \to Linear \to ELU \to Linear \to ELU \to \cdots$.
The policy network had three layers of [256, 256, 256] units respectively. Actions are sampled from a multivariate Gaussian with diagonal covariance, parameterized by initial and minimum scale parameters for which we use the default values of 0.3 and $10^{-6}$ respectively. The critic network had three layers of [512, 512, 256] units, the output of the last layer is concatenated with the actions and passed into the discrete valued head that returns the categorical distribution, using 51 categories spaced evenly across [-15, 15].

\paragraph{DDPG}

Just like in case of DMPO both policy and critic networks are LayerNorm MLPs where we use the same hyperparameters. The policy network is the LayerNorm MLP with [256, 256, 256] units. The critic network concatenates observations and actions and passes them into the LayerNorm MLP with [512, 512, 256] units, which go into another Linear Layer with output dimension 1.

\paragraph{D4PG}

For D4PG we use the critic network from DDPG and the policy networks from DMPO with the same hyperparameters.

\section{Limitations}

Some limitations of our model and the Industrial Task Suite include:
\begin{itemize}
    \item \textit{Simulation Speed}: The simulation software solves complex ordinary differential equations at every step. Even with our hybrid analytical model, this results in a magnitude of $\sim10$ seconds/step with each step simulating a real world 5 minute time horizon. With the current setup, this doesn't match the high frequency iterations of existing reinforcement learning environments. A next step would involve optimizing the environment setup and the solver for faster simulated transitions.
    
    \item \textit{Simulation Fidelity} The simulator fidelity checks described in this paper are not exhaustive; more work is required to ensure its accuracy.
    
    \item \textit{Sim2Real Validation}:  We have designed and implemented the Industrial Task Suite to enable domain randomization, a technique that randomizes the simulator parameters to expose the agent to a diverse set of environments in the training phase. Additionally, it provides access to a wide diversity of real world scenarios. However, it remains a future work to develop a Sim2real adapter to integrate this with a real world facility. 
\end{itemize}

The modular Industrial Task Suite, as well as the hybrid simulation model design, have been described in detail. While this should be descriptive to provide a high level understanding of the contribution, releasing the codebase remains out of scope for this work.

\section{Conclusions}

Designing and evaluating accurate whole-facility industrial cooling models is a challenging optimization problem. In this paper, we presented one possible approach by combining analytical solutions with a multi-physics simulation software. We showed that it empirically matches key properties of real-world facility behavior.

We extended and demonstrated the usability of this model by providing the design for an Industrial Task Suite, which allows custom RL problem configurations using our model or any other physical control simulator. We benchmarked RL agents using this task suite and identified DMPO to be a well-performing candidate for the problem of cooling system control, as it found the optimal policies for all four given tasks. Providing this model and framework, we hope to make applying RL to the real-world problem of optimizing industrial controls more accessible.

\newpage

\appendix

\begin{center}
{\Large \bf Appendix}
\end{center}

\section{Industrial Task Suite Components}
\label{sec:its-components}

In this section we will briefly describe all components of the task suite defined in Section \ref{sec:its} and provide some implementation details.

\subsection{Environment}

\texttt{Environment} extends \texttt{dm\_env.Environment}, an abstract base class for RL environments defined in The DeepMind RL Environment API \cite{dm_env2019}. Its main job is to comprise and orchestrate all components (see Figure \ref{fig:its-design}).
\begin{lstlisting}[language=Python]
class Environment(dm_env.Environment):
  def __init__(self,
               simulator: SimulatorInterface,
               task: TaskInterface,
               evaluator: Evaluator,
               initial_conditions: Mapping[str, Any],
               initial_conditions_noise: Mapping[str,float],
               controls: Collection[str],
               controls_noise: Mapping[str, float],
               measurements: Collection[str],
               measurements_noise: Mapping[str, float]
               ...)
    self._simulator = simulator
    self._task = task
    self._evaluator = evaluator
    ...
    self._ic = initial_conditions
    self._ic_noise = initial_conditions_noise
    ...
    self._controls_noise = controls_noise
    ...

  @property
  def action_spec(self):

  @property
  def observation_spec(self):

  def reset(self):
    # Add noise to initial conditions.
    noisy_cfg = {}
    for ic_id, ic_value in self._ic.items():
      noisy_cfg[ic_id] = ic_value + self._ic_noise[ic_id]

    # Reset the simulator with the new noisy initial conditions
    timestep = self._simulator.reset(noisy_cfg)
    
    # Add noise to measurements and evaluate constraints
    core_obs, _ = process_timestep(timestep, self._evaluator)
    
    # Reset task.
    task_obs = self._task.reset(noisy_cfg, core_obs)
    
    observation = dict(**core_obs, **task_obs)
    return dm_env.restart(observation)

  def step(self, action):
    # Add noise to controls (actions)
    for action_id in action:
      action[action_id] += self._controls_noise[action_id]
      
    timestep = self._simulator.step(action)
    
    # Add noise to measurements and evaluate constraints
    core_obs, hard_constraint_violation = process_timestep(timestep, self._evaluator)
    
    # Compute reward and extend observation
    reward, task_obs = self._task.calculate_reward_and_observations(
        core_obs)
    
    observation = dict(**core_obs, **task_obs)
    if timestep.is_last or hard_constraint_violation:
      return dm_env.termination(reward, observation)
    else:
      return dm_env.transition(reward, observation)
\end{lstlisting}

\subsection{Tasks}

Tasks are designed to exercise orthogonal areas of competence, but at the same time work with any configuration sampled from the defined distribution.  In addition to that,  the shared  component  framework  allows  testing out certain policies on the entire range of tasks. From the implementation point of view tasks define and compute rewards as well as provide additional observations to the agent.

\begin{lstlisting}[language=Python]
class TaskInterface:

  @property
  def observation_spec(self):

  def reset(self, configuration):

  def calculate_reward_and_observations(observation):
\end{lstlisting}

\subsection{Simulator}

A simulator is the ground-truth model of the system dynamics. It is configured at the beginning of each new episode. That is the moment when a new configuration will be sampled. This means that at each new episode, a new simulation might be selected as well. Once the simulator is configured it starts taking controls from the agent at every agent step and returns the results in the form of measurements, which are a subset of the simulation state.

\begin{lstlisting}[language=Python]
class SimulatorInterface:

  @property
  def measurement_spec(self):

  def reset(self, configuration):

  def step(self, controls):
\end{lstlisting}

\subsection{Constraints}
\label{sec:constraints}

\textit{Constraints} are defined in a configuration file. In the industrial setting the agent gets kicked out of the system, when it violates some constraints, to mimic this we divide constraints into hard and soft. Violation of a hard constraint automatically ends an episode (equivalent to the agent being kicked out of the system). Violation of any of these constraints is communicated to the agent as an observation.
Constraints are evaluated on every step - after receiving an agent’s action, and subsequently after receiving an updated simulation measurements. The constraint evaluator is used in both cases to validate them.

\begin{figure}[h!]
     \includegraphics[width=\columnwidth]{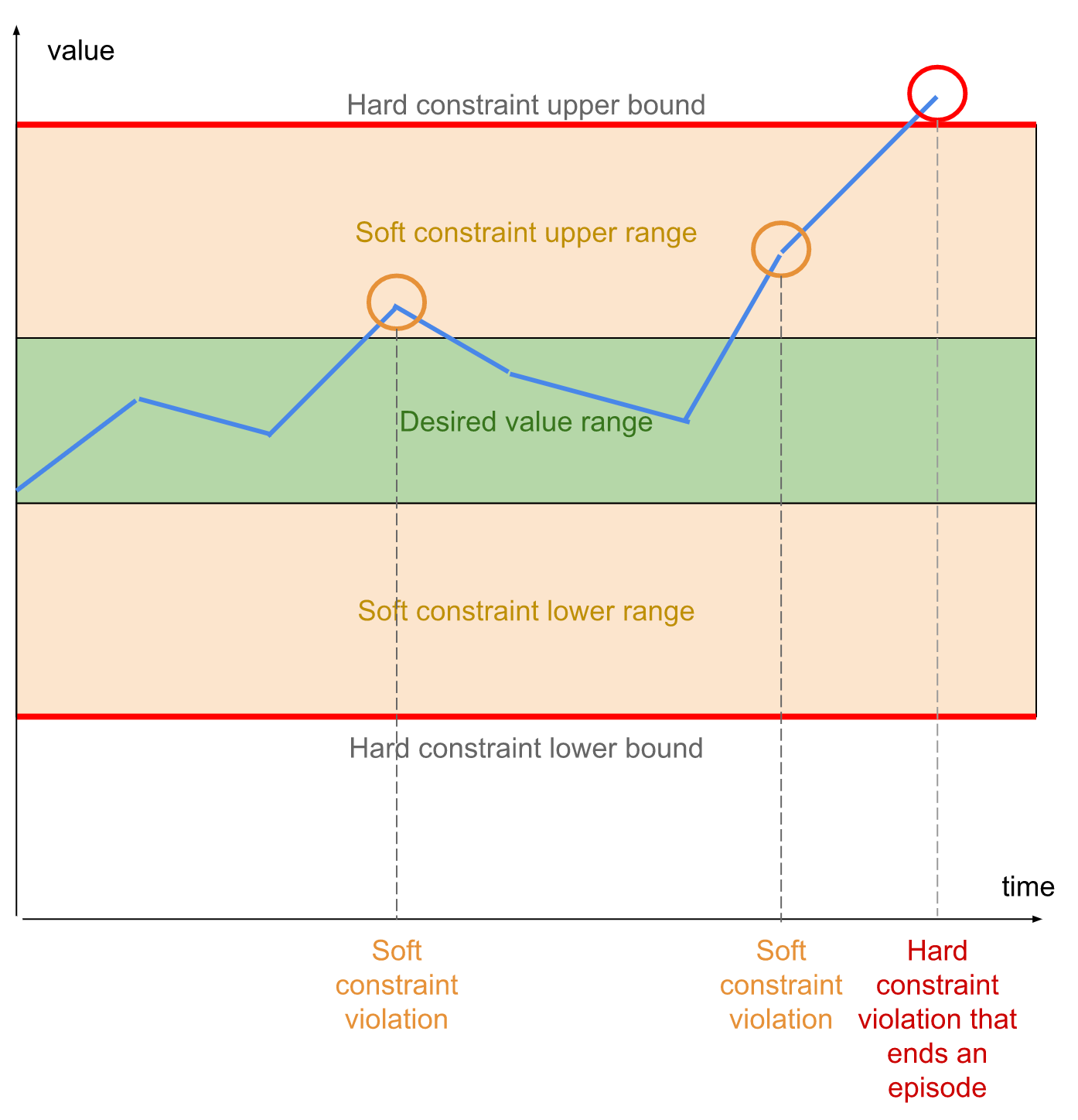}
    \caption{Evaluation of the constraint ranges for a single value.}
    \label{fig:constraints}
\end{figure}

Hard constraints can be expressed by a single range of values. Soft constraints define ranges bounded by the hard constraint upper and lower limits. This can be expressed with a custom structure that defines four values for each of the limits.

\begin{lstlisting}[language=Python]
class Constraint:
  simulation_parameter_name: str
  hard_upper: float
  soft_upper: float
  soft_lower: float
  hard_lower: float
\end{lstlisting}

The values should meet the following criteria
\texttt{hard\_lower <= soft\_lower < soft\_upper <= hard\_upper}.

\subsection{Scenarios}

Scenarios add another configuration dimension to the environment. Each scenario describes a trajectory that modifies a set of simulation configuration parameters. Along with noise, scenarios make environment more realistic. Below we list a set of scenarios each representing a challenge we observed in the real world industrial systems.

\begin{itemize}

\item Frozen sensors: The scenario freezes the values of selected sensors for a random amount of time. A dedicated implementation of the measurement noise handles this. Its purpose is to test an agent’s resilience to missing information.

\item Sensor drift: The scenario introduces a temporally correlated noise into the selected measurement components. The components are selected at random at the beginning of each episode. Its purpose is to test an agent’s resilience to partially false information.

\item Frozen controls: The scenario freezes the values of selected controls for a random amount of time. A dedicated implementation of the control noise handles this.Its purpose is to test an agent's ability to detect when a selected policy fails and adapt by switching to an alternative.

\item Simulation dynamics non-stationarity: The scenario uses a set of configuration trajectories to modify the selected simulation configuration parameters over the course of an episode. Configuration trajectories produce changes to selected parameters, which the task adds to their baseline values and subsequently passes to the simulation. Control tensor is used for that purpose.  Absolute configuration values are used so that limits can be observed. The limits are stored in the configuration definition. Its purpose is to test an agent's resilience to ever changing environmental conditions, for example, weather change and other variables outside the domain of agent’s control.

\end{itemize}

\section{Other Industrial Task Suite Applications}
\label{sec:its-other-applications}

The modularity of the industrial task suite and the tools developed around the cooling system simulation model allow the easy implementation of other control problems of interest, which might have very different physics and task definitions.

\subsection{Brushless DC motor}
Direct control electrical machines is a long standing research problem for industrial control systems, and using reinforcement learning based control approaches offers potential to improve on the state-of-the-art performance of existing controllers. 

\begin{figure*}[h]
    \includegraphics[width=\textwidth]{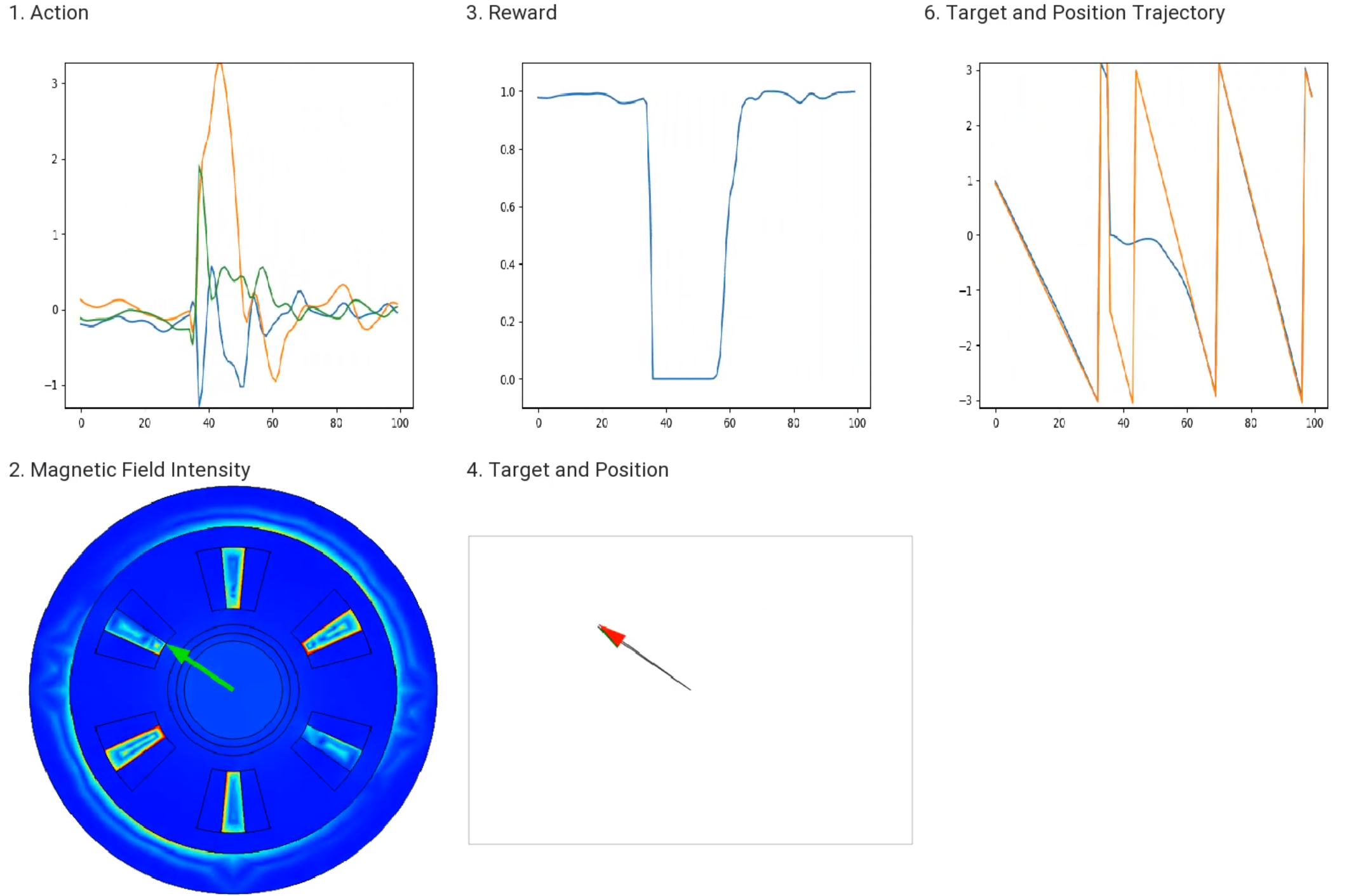}
    \caption{BLDC motor simulation environment captured as stills from video \cite{motor-control-video}. The agent sets actions in the form of three phase currents (top left), to control the motor. (Bottom Center) Target position is shown with a red arrow and the current position is shown with a green arrow.}
    \label{fig:bldc_motor_full_plot}
\end{figure*}

In order to explore research into this area, we have implemented a simulation model for a Brushless DC (BLDC) electrical motor using COMSOL. The simulation model is a 2D axial cross-section of a 3-phase electrical motor, with a magnetized rotary core (see Figure ~\ref{fig:bldc_motor_full_plot}). The agent observes the angular position and velocity of the rotor, and controls the currents being applied to each of the three phase coils. The simulation also allows us to visualize (Figure~\ref{fig:bldc_motor_full_plot}-2) the magnetic field intensity of the simulated motor.

We have successfully trained a DMPO agent \cite{abdolmaleki2018maximum, acme-DMPO-agent} to control the motor in a generic trajectory tracking task. A random target position trajectory is created and sent to the agent; and a reward is calculated based on the difference between the actual and target positions, with reward = 1 when they are equal. The agent also uses a secondary objective to reduce the size of actions ("phase currents"), to minimize power.  

\subsection{Airfoil}
Another example domain with complex physics is fluid flow. Flight control systems aim to achieve lift and stability while countering non-linear and time-varying physics~\cite{kim1997nonlinear}. This domain provides a challenging test domain for our agents and our methods for providing and soft/hard constraints to achieve stability. 

We have implemented an airfoil under flow to control the altitude. The agent observes the current altitude, a target altitude and current pitch of the airfoil in addition to the lift force generated, and controls the rate of change of the pitch. We have implemented a DMPO agent to learn a control policy that achieves a target altitude. We have successfully demonstrated this policy in the following video~\cite{airfoil-control-video}.   

\bibliography{bibliography}

\begin{thebibliography}{10}

\bibitem{cooling-system-britannica}
Britannica, t. editors of encyclopaedia (2016, february 16). cooling system.
  encyclopedia britannica.
\newblock https://www.britannica.com/technology/cooling-system.

\bibitem{google-ai-data-center}
Deepmind ai reduces energy used for cooling google data centers by 40
\newblock
  \url{https://blog.google/outreach-initiatives/environment/deepmind-ai-reduces-energy-used-for/}.
\newblock Accessed: 2022-07-26.

\bibitem{acme-DMPO-agent}
Distributional maximum a posteriori policy optimization (dmpo).
\newblock
  \url{https://github.com/deepmind/acme/tree/master/acme/agents/tf/dmpo}.
\newblock Accessed: 2022-07-11.

\bibitem{airfoil-control-video}
Dmpo agent learns to control a airfoil under fluid flow.
\newblock \url{https://youtu.be/fGRDnwYYJnw}.
\newblock Accessed: 2022-07-12.

\bibitem{motor-control-video}
Dmpo agent learns to control a bldc motor to achieve "constant speed" task.
\newblock \url{https://youtu.be/dR2cT8cRTlw}.
\newblock Accessed: 2022-07-12.

\bibitem{industrial-sector-energy-consumption}
Industrial sector energy consumption.
\newblock https://www.eia.gov/outlooks/ieo/pdf/industrial.pdf.

\bibitem{abdolmaleki2018maximum}
A.~Abdolmaleki, J.~T. Springenberg, Y.~Tassa, R.~Munos, N.~Heess, and
  M.~Riedmiller.
\newblock Maximum a posteriori policy optimisation.
\newblock {\em arXiv preprint arXiv:1806.06920}, 2018.

\bibitem{barth2018distributed}
G.~Barth-Maron, M.~W. Hoffman, D.~Budden, W.~Dabney, D.~Horgan, D.~Tb,
  A.~Muldal, N.~Heess, and T.~Lillicrap.
\newblock Distributed distributional deterministic policy gradients.
\newblock {\em arXiv preprint arXiv:1804.08617}, 2018.

\bibitem{bengio2009curriculum}
Y.~Bengio, J.~Louradour, R.~Collobert, and J.~Weston.
\newblock Curriculum learning.
\newblock In {\em Proceedings of the 26th annual international conference on
  machine learning}, pages 41--48, 2009.

\bibitem{brockman2016openai}
G.~Brockman, V.~Cheung, L.~Pettersson, J.~Schneider, J.~Schulman, J.~Tang, and
  W.~Zaremba.
\newblock Openai gym.
\newblock {\em arXiv preprint arXiv:1606.01540}, 2016.

\bibitem{cao2018optimal}
Y.~Cao, L.~Huang, Z.~Cui, and J.~Liu.
\newblock The optimal operation of cooling tower systems with
  variable-frequency control.
\newblock In {\em IOP Conference Series: Earth and Environmental Science},
  volume 113, page 012085. IOP Publishing, 2018.

\bibitem{chen2019gnu}
B.~Chen, Z.~Cai, and M.~Berg{\'e}s.
\newblock Gnu-rl: A precocial reinforcement learning solution for building hvac
  control using a differentiable mpc policy.
\newblock In {\em Proceedings of the 6th ACM international conference on
  systems for energy-efficient buildings, cities, and transportation}, pages
  316--325, 2019.

\bibitem{chen2018optimal}
Y.~Chen, Y.~Shi, and B.~Zhang.
\newblock Optimal control via neural networks: A convex approach.
\newblock {\em arXiv preprint arXiv:1805.11835}, 2018.

\bibitem{crawley2001energyplus}
D.~B. Crawley, L.~K. Lawrie, F.~C. Winkelmann, W.~F. Buhl, Y.~J. Huang, C.~O.
  Pedersen, R.~K. Strand, R.~J. Liesen, D.~E. Fisher, M.~J. Witte, et~al.
\newblock Energyplus: creating a new-generation building energy simulation
  program.
\newblock {\em Energy and buildings}, 33(4):319--331, 2001.

\bibitem{dulac2019challenges}
G.~Dulac-Arnold, D.~Mankowitz, and T.~Hester.
\newblock Challenges of real-world reinforcement learning.
\newblock {\em arXiv preprint arXiv:1904.12901}, 2019.

\bibitem{foliaco2020improving}
B.~Foliaco, A.~Bula, and P.~Coombes.
\newblock Improving the gordon-ng model and analyzing thermodynamic parameters
  to evaluate performance in a water-cooled centrifugal chiller.
\newblock {\em Energies}, 13(9):2135, 2020.

\bibitem{gao2019energy}
G.~Gao, J.~Li, and Y.~Wen.
\newblock Energy-efficient thermal comfort control in smart buildings via deep
  reinforcement learning.
\newblock {\em arXiv preprint arXiv:1901.04693}, 2019.

\bibitem{gordon1995centrifugal}
J.~Gordon, K.~C. Ng, and H.~T. Chua.
\newblock Centrifugal chillers: thermodynamic modelling and a diagnostic case
  study.
\newblock {\em International Journal of refrigeration}, 18(4):253--257, 1995.

\bibitem{hanif2022renewable}
M.~A. Hanif, F.~Nadeem, R.~Tariq, and U.~Rashid.
\newblock {\em Renewable and Alternative Energy Resources}.
\newblock Academic Press, 2022.

\bibitem{hanumaiah2021distributed}
V.~Hanumaiah and S.~Genc.
\newblock Distributed multi-agent deep reinforcement learning framework for
  whole-building hvac control.
\newblock {\em arXiv preprint arXiv:2110.13450}, 2021.

\bibitem{hertwich2019material}
E.~G. Hertwich, S.~Ali, L.~Ciacci, T.~Fishman, N.~Heeren, E.~Masanet, F.~N.
  Asghari, E.~Olivetti, S.~Pauliuk, Q.~Tu, et~al.
\newblock Material efficiency strategies to reducing greenhouse gas emissions
  associated with buildings, vehicles, and electronics—a review.
\newblock {\em Environmental Research Letters}, 14(4):043004, 2019.

\bibitem{hoffman2020acme}
M.~Hoffman, B.~Shahriari, J.~Aslanides, G.~Barth-Maron, F.~Behbahani,
  T.~Norman, A.~Abdolmaleki, A.~Cassirer, F.~Yang, K.~Baumli, S.~Henderson,
  A.~Novikov, S.~G. Colmenarejo, S.~Cabi, C.~Gulcehre, T.~L. Paine, A.~Cowie,
  Z.~Wang, B.~Piot, and N.~de~Freitas.
\newblock Acme: A research framework for distributed reinforcement learning.
\newblock {\em arXiv preprint arXiv:2006.00979}, 2020.

\bibitem{Kazmi_2018}
H.~Kazmi, F.~Mehmood, S.~Lodeweyckx, and J.~Driesen.
\newblock Gigawatt-hour scale savings on a budget of zero: Deep reinforcement
  learning based optimal control of hot water systems.
\newblock {\em Energy}, 144:159--168, feb 2018.

\bibitem{kim1997nonlinear}
B.~S. Kim and A.~J. Calise.
\newblock Nonlinear flight control using neural networks.
\newblock {\em Journal of Guidance, Control, and Dynamics}, 20(1):26--33, 1997.

\bibitem{kim2011difficulties}
D.-W. Kim and C.-S. Park.
\newblock Difficulties and limitations in performance simulation of a double
  skin fa{\c{c}}ade with energyplus.
\newblock {\em Energy and Buildings}, 43(12):3635--3645, 2011.

\bibitem{krausmann2018resource}
F.~Krausmann, C.~Lauk, W.~Haas, and D.~Wiedenhofer.
\newblock From resource extraction to outflows of wastes and emissions: The
  socioeconomic metabolism of the global economy, 1900--2015.
\newblock {\em Global Environmental Change}, 52:131--140, 2018.

\bibitem{liang2018rllib}
E.~Liang, R.~Liaw, R.~Nishihara, P.~Moritz, R.~Fox, K.~Goldberg, J.~Gonzalez,
  M.~Jordan, and I.~Stoica.
\newblock Rllib: Abstractions for distributed reinforcement learning.
\newblock In {\em International Conference on Machine Learning}, pages
  3053--3062. PMLR, 2018.

\bibitem{lillicrap2015continuous}
T.~P. Lillicrap, J.~J. Hunt, A.~Pritzel, N.~Heess, T.~Erez, Y.~Tassa,
  D.~Silver, and D.~Wierstra.
\newblock Continuous control with deep reinforcement learning.
\newblock {\em arXiv preprint arXiv:1509.02971}, 2015.

\bibitem{mnih2016asynchronous}
V.~Mnih, A.~P. Badia, M.~Mirza, A.~Graves, T.~Lillicrap, T.~Harley, D.~Silver,
  and K.~Kavukcuoglu.
\newblock Asynchronous methods for deep reinforcement learning.
\newblock In {\em International conference on machine learning}, pages
  1928--1937. PMLR, 2016.

\bibitem{moriyama2018reinforcement}
T.~Moriyama, G.~D. Magistris, M.~Tatsubori, T.-H. Pham, A.~Munawar, and
  R.~Tachibana.
\newblock Reinforcement learning testbed for power-consumption optimization.
\newblock In {\em Asian simulation conference}, pages 45--59. Springer, 2018.

\bibitem{dm_env2019}
A.~Muldal, Y.~Doron, J.~Aslanides, T.~Harley, T.~Ward, and S.~Liu.
\newblock dm\_env: A python interface for reinforcement learning environments,
  2019.

\bibitem{mun2020limitations}
S.-H. Mun, J.~Kang, Y.~Kwak, Y.-S. Jeong, S.-M. Lee, and J.-H. Huh.
\newblock Limitations of energyplus in analyzing energy performance of
  semi-transparent photovoltaic modules.
\newblock {\em Case Studies in Thermal Engineering}, 22:100765, 2020.

\bibitem{nagy2021real}
Z.~Nagy and K.~Nweye.
\newblock Real-world challenges for reinforcement learning in building control.
\newblock {\em arXiv preprint arXiv:2112.06127}, 2021.

\bibitem{narvekar2020curriculum}
S.~Narvekar, B.~Peng, M.~Leonetti, J.~Sinapov, M.~E. Taylor, and P.~Stone.
\newblock Curriculum learning for reinforcement learning domains: A framework
  and survey.
\newblock {\em arXiv preprint arXiv:2003.04960}, 2020.

\bibitem{portner2022climate}
H.~O. Portner, D.~C. Roberts, H.~Adams, C.~Adler, P.~Aldunce, E.~Ali, R.~A.
  Begum, R.~Betts, R.~B. Kerr, R.~Biesbroek, et~al.
\newblock Climate change 2022: impacts, adaptation and vulnerability.
\newblock 2022.

\bibitem{tassa2018deepmind}
Y.~Tassa, Y.~Doron, A.~Muldal, T.~Erez, Y.~Li, D.~d.~L. Casas, D.~Budden,
  A.~Abdolmaleki, J.~Merel, A.~Lefrancq, et~al.
\newblock Deepmind control suite.
\newblock {\em arXiv preprint arXiv:1801.00690}, 2018.

\bibitem{tortorelli2022parallel}
A.~Tortorelli, M.~Imran, F.~Delli~Priscoli, and F.~Liberati.
\newblock A parallel deep reinforcement learning framework for controlling
  industrial assembly lines.
\newblock {\em Electronics}, 11(4):539, 2022.

\bibitem{wang2019evolution}
J.~Wang, J.~F. Rodrigues, M.~Hu, P.~Behrens, and A.~Tukker.
\newblock The evolution of chinese industrial co2 emissions 2000--2050: A
  review and meta-analysis of historical drivers, projections and policy goals.
\newblock {\em Renewable and Sustainable Energy Reviews}, 116:109433, 2019.

\bibitem{zhan2021deepthermal}
X.~Zhan, H.~Xu, Y.~Zhang, Y.~Huo, X.~Zhu, H.~Yin, and Y.~Zheng.
\newblock Deepthermal: Combustion optimization for thermal power generating
  units using offline reinforcement learning.
\newblock {\em arXiv preprint arXiv:2102.11492}, 2, 2021.

\bibitem{zhang2019whole}
Z.~Zhang, A.~Chong, Y.~Pan, C.~Zhang, and K.~P. Lam.
\newblock Whole building energy model for hvac optimal control: A practical
  framework based on deep reinforcement learning.
\newblock {\em Energy and Buildings}, 199:472--490, 2019.

\end{thebibliography}
\end{document}